\documentclass[journal,twoside]{IEEEtran}
\IEEEoverridecommandlockouts
\usepackage{cite}
\usepackage{amsmath,amssymb,amsfonts}
\usepackage{graphicx}
\usepackage{textcomp}
\usepackage{xcolor}

\usepackage[ruled]{algorithm2e}
\usepackage{subfigure}
\usepackage{bm}
\usepackage{amssymb}
\usepackage{threeparttable}
\usepackage{booktabs}
\usepackage{multirow}
\def\BibTeX{{\rm B\kern-.05em{\sc i\kern-.025em b}\kern-.08em
    T\kern-.1667em\lower.7ex\hbox{E}\kern-.125emX}}
\begin{document}
	
\title{What and Where: Learn to Plug Adapters via NAS for Multi-Domain Learning}

\author{Hanbin Zhao, Hao Zeng, Xin Qin, Yongjian Fu, Hui Wang, Bourahla Omar, Xi Li*
	
	\thanks{H. Zhao,  H. Zeng, Q. Xin, Y. Fu, H. Wang, O. Bourahla are with College of Computer Science, Zhejiang University, Hangzhou 310027, China. (e-mail:
		{zhaohanbin, zenghao\_97, xinqin, yjfu, wanghui\_17, bourahla@zju.edu.cn})}%
	\thanks{X. Li* (corresponding author) is with the College of Computer Science and Technology, Zhejiang University, Hangzhou 310027, China and also with the Shanghai Institute for Advanced Study, Zhejiang University, Shanghai 201210, China (e-mail: {xilizju@zju.edu.cn} phone: 0571-87951247)}%
}
\markboth{IEEE Transactions on Neural Networks and Learning Systems,~Vol.~XX, No.~X, 2021}%
{Zhao \MakeLowercase{\textit{et al.}}: What and Where: Learn to Plug Adapters via NAS for Multi-Domain Learning}
\maketitle

\begin{abstract}
As an important and challenging problem, multi-domain learning (MDL) typically seeks for a set of effective lightweight domain-specific adapter modules
plugged into a common domain-agnostic network. Usually, existing ways of adapter plugging and structure design are handcrafted and fixed for all domains before model learning, resulting in the learning inflexibility and computational intensiveness. With this motivation, we propose to learn a data-driven adapter plugging strategy with Neural Architecture Search (NAS), which automatically determines where to plug for those adapter modules. Furthermore, we propose a NAS-adapter module for adapter structure design in a NAS-driven learning scheme, which automatically discovers effective adapter module structures for different domains. Experimental results demonstrate the effectiveness of our MDL model against existing approaches under the conditions of comparable performance. 
\end{abstract}
\begin{IEEEkeywords}
Multi-Domain Learning, Adapter, Image Classification, Neural Architecture Search
\end{IEEEkeywords}

\section{Introduction}\label{sec:introduction}
\IEEEPARstart{R}ecent years have witnessed a great development of Convolutional Neural Networks (CNNs) together with a wide variety of their vision applications. For the sake of high performance, these networks devote great efforts to carefully designing complicated structures for different tasks in a domain-specific manner, leading to the
inflexibility of model learning across multiple domains. 
As a result, in the case of several tasks based on different domains, one needs to deploy an equal number of domain-specific models respectively, which is unrealizable in practice, especially when concerning the limitation of computational resources.
To tackle the problem, \textit{Multi-domain} learning~\cite{rebuffi2017learning,rebuffi2018efficient,li2019efficient, liu2018multi, yang2019multi} emerges as an important approach for better efficiency and generalization of model learning across multiple different yet correlated domains.

\begin{figure}[t]
	\centering
	\includegraphics[width=1\columnwidth]{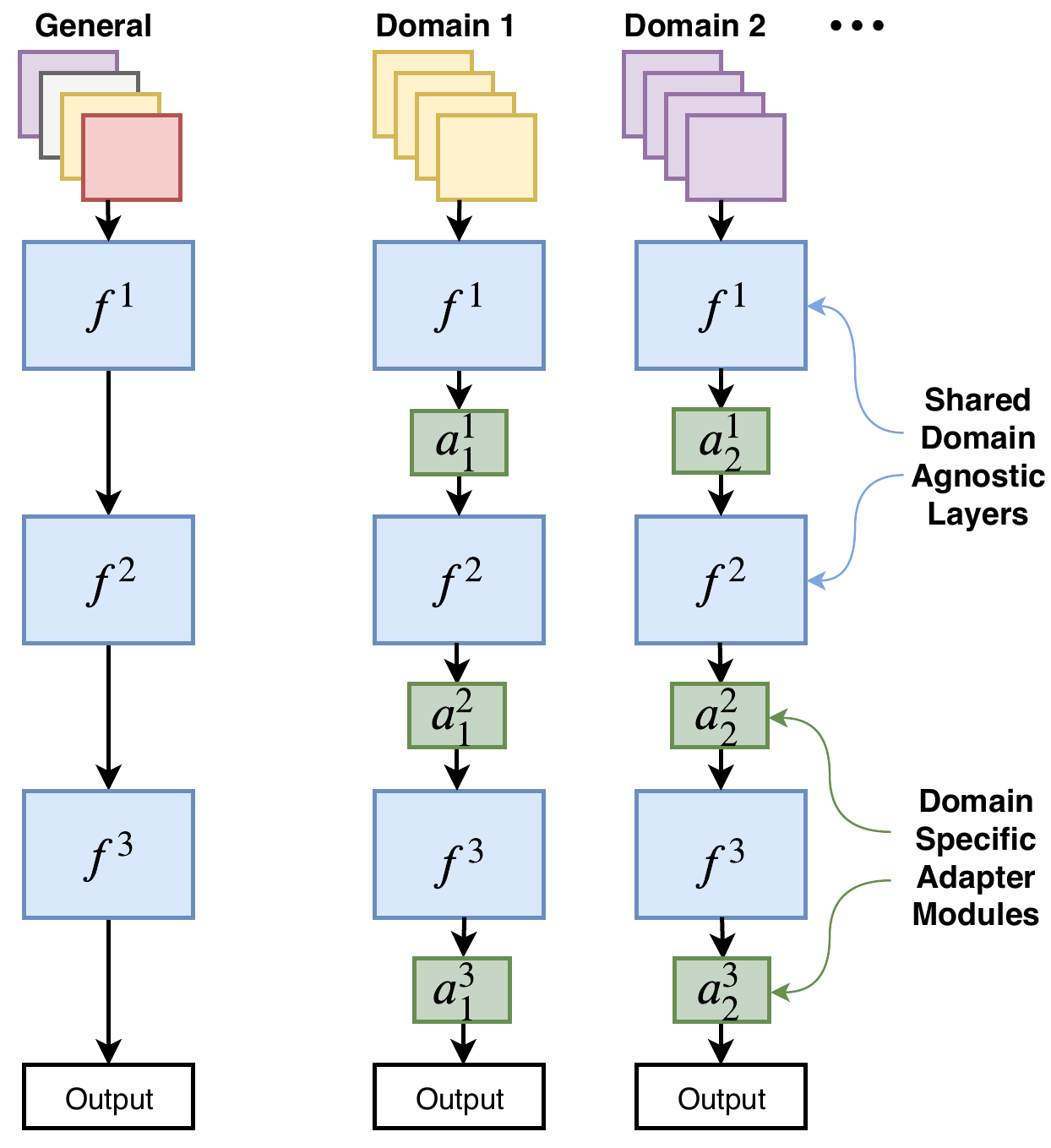}
	\caption{Illustration of \textit{multi-domain} learning. The first stage is to seek a \textit{domain-agnostic} model (e.g. VGG) as our common trunk model, and the second is to plug a set of \textit{domain-specific} adapter modules into the former, leading to the final adaptation model. After learning the adapter parameters with the parameters of the trunk model fixed, each domain can be adapted by changing a set of adapter modules.}
	\label{fig:multi_domain_learning}
\end{figure}

In principle, \textit{multi-domain learning} (MDL)~\cite{rebuffi2017learning} aims to learn a compact model that works well for many different domains (e.g., internet images, scene text, medical images, satellite images, driving images, etc.). Typically, it is cast as a two-stage learning problem (illustrated in Figure~\ref{fig:multi_domain_learning}), including domain-agnostic model learning and domain-specific model adaptation. Specifically, domain-agnostic model learning is to seek a common trunk neural network model (with structures and parameters shared across domains). In comparison, domain-specific model adaptation aims at plugging a set of extremely lightweight adapter modules into the common trunk model structure for dynamically adapting to different domains. After learning the adapter parameters with
the common model fixed, we have domain-specific models that are pretty flexible in terms of changing the adapter modules.
In sum, the core content of MDL is to design a plugging strategy (i.e., where to plug) as well as a set of adapter module structures (i.e. what to plug), which directly determines the effectiveness and the compactness of the whole adaptation model.

In the research context, the adapter plugging strategy~\cite{berriel2019budget,rebuffi2017learning,rebuffi2018efficient,li2019efficient, bulat2019incremental} is usually fixed, dense, and handcrafted. Consequently, it is less flexible and discriminative with a higher computational cost in many complicated situations. Moreover, the adapter module structure is also predefined and fixed for different domains, leading to the weakness in cross-domain adaptation. Therefore, how to automatically set up the adapter plugging strategy
and adaptively fulfill the adapter structure design are crucial to effective multi-domain learning.
Motivated by this observation, we propose a novel NAS-driven scheme for multi-domain learning based on Neural Architecture Search (NAS). Specifically,
we accomplish the task of automatically finding the effective adapter plugging strategy by NAS, and meanwhile make full use of NAS to search the adapter structures adaptively. In this way, our scheme has the following advantages: 1) more flexible and sparse adapter plugging with better efficiency and generalization by NAS; and 2) more discriminative adapter modules with better adaptation to different domains. As a result, the multi-domain model we obtain is often more compact, discriminative, and domain adaptive with a relatively low computational cost when compared to previous MDL methods.

In summary, the main contributions of this work are summarized as follows:
\begin{itemize}
	\item We propose a NAS-driven scheme for multi-domain learning, which effectively makes model learning seamlessly adapt to different domains by automatically determining where to plug with NAS.
	\item We propose a NAS-adapter module which adaptively discovers the adapter module structures by NAS for well balancing between the model effectiveness and compactness for different domains.
	\item Extensive experiments over benchmark datasets demonstrate the effectiveness of this work in accuracy and flexibility against the existing approaches.	
\end{itemize}

The rest of the paper is organized as follows. We first describe the background in Section~\ref{related_work}, and then explain the details of our proposed scheme in Section~\ref{method}. In Section~\ref{experiments}, we conduct the experiments and discuss their corresponding results. Finally, we conclude this work in Section~\ref{conclusion}.

\begin{figure*}[t]
	\centering
	\includegraphics[width=1\textwidth]{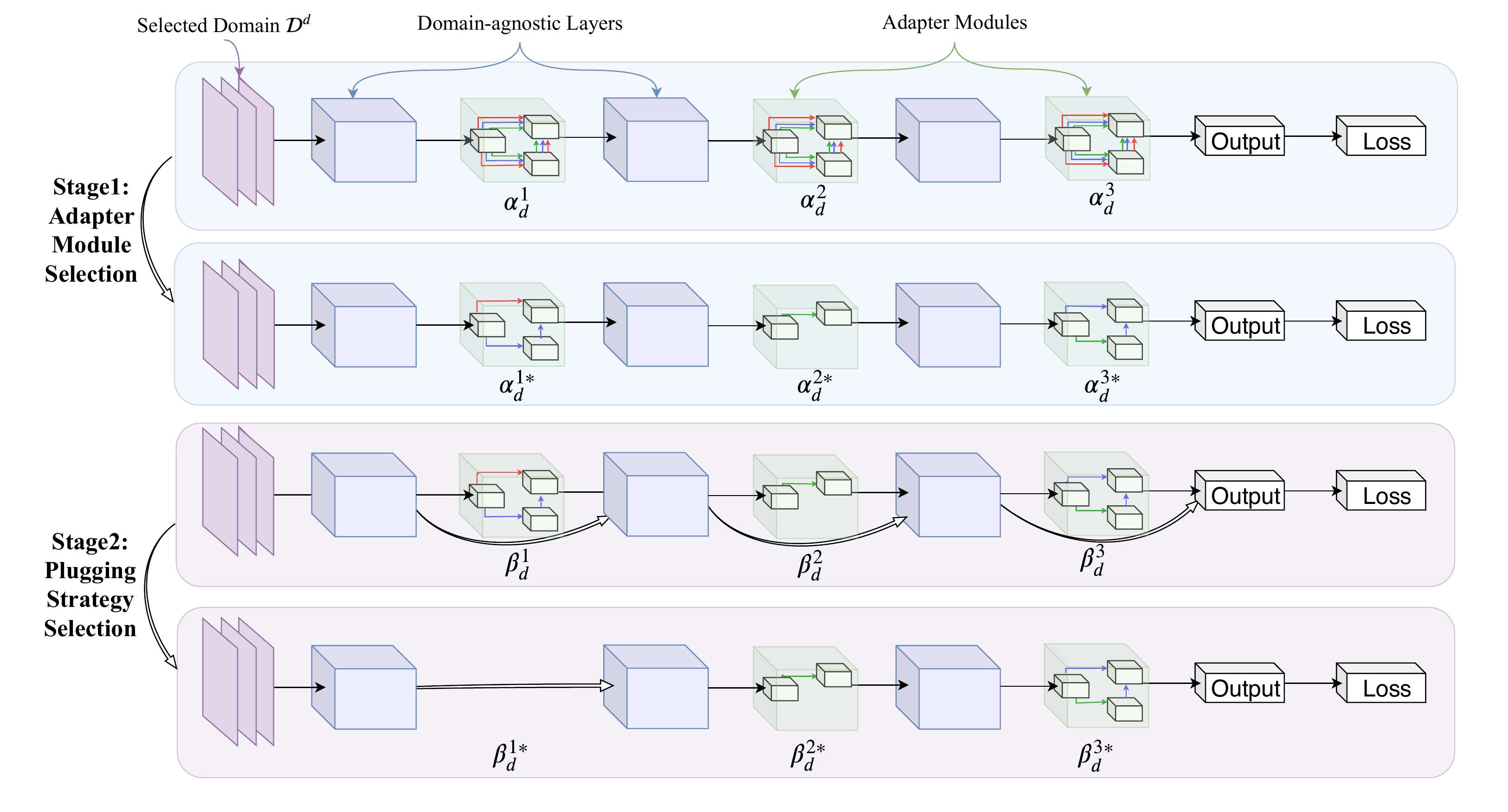}
	\caption{Illustration of our scheme for multi-domain learning. In the first stage, we search a set of appropriate adapters according to the given domain and the plugging location. In the second stage, we select an adapter plugging strategy (i.e. where to plug the adapter) to further compact the adaptation model.}
	\label{fig:our_method}
\end{figure*}

\section{Related Work}\label{related_work}
\subsection{Multi-Domain Learning} 
MDL~\cite{rebuffi2017learning,rebuffi2018efficient,li2019efficient, liu2018multi, yang2019multi, gu2018multi, yang2019shared,liu2019compact, fourure2017multi, guo2019depthwise} aims to learn a model that works well for several different visual domains, requiring both effectiveness and efficiency.
Various adapter modules are proposed to gain an acceptable performance, overcoming the ``catastrophic forgetting''~\cite{mccloskey1989catastrophic,pfulb2019comprehensive,coop2013ensemble} problem. 
BN~\cite{bilen2017universal} used the batch-normalization layer as the domain-specific adapter and therefore only a few parameters are finetuned for each domain. 
To tackle the complex visual domains, more powerful adapter modules are proposed, such as the $1 \times 1$ convolutional adapter~\cite{rosenfeld2018incremental} and the residual adapter (RA)~\cite{rebuffi2017learning}, which gain significant performance while increasing the computational resource cost in exchange.
Those methods all have one thing in common, that the adapters are all hand-crafted, and the same adapter structure is used for all different domains. 
In this paper, we claim that the structure of adapters should also be adapted, along with the change of domains.
Our method learns the structure of adapters while taking the domain diversity and possible plugging locations into consideration.

On the other hand, some of the MDL methods pay more attention to compress the parameters further thereby improving efficiency. RA-SVD~\cite{rebuffi2018efficient} proposed to compress the adapter with the singular matrix decomposition (SVD). CovNorm~\cite{li2019efficient} utilized principal component analysis (PCA) aligned by the covariance from data to compress the adapter modules. While these methods concentrate on the compression within the adapter modules, we proposed to further compact the whole adaptation model in a domain-specific fashion. Adapters are plugged only into several selected locations, and the plugging strategy varies from different domains. Computational resources therefore can be obviously saved without performance decreasing.


\subsection{Neural Architecture Search} 
NAS~\cite{elsken2018neural,cai2018proxylessnas,kandasamy2018neural} aims at designing effective neural network architectures automatically. There is a rich body of works in NAS, which are mainly based on three strategies: reinforcement learning method~\cite{baker2016designing,zoph2016neural,zhang2019customizable,tan2019mnasnet}, evolutionary algorithm~\cite{real2017large}, and gradient-based optimization~\cite{liu2018darts, chen2019progressive, xu2019pc,vahdat2019unas,jin2019rc}. Because of its remarkable performance, a lot of NAS-based methods have been proposed to solve some specific problems. Auto-DeepLab~\cite{liu2019auto} proposed a hierarchical search space for semantic segmentation task~\cite{feng2020taplab,chen2020banet,sun2021real,ji2019human,ji2020context}. FP~\cite{newell2019feature} proposed a searching strategy for designing efficient multi-task architecture. In life-long learning~\cite{aljundi2018memory,kirkpatrick2017overcoming,li2017learning,zhao2021memory,zhao2006mgsvf}, LTG~\cite{li2019learn} proposed to expand the network architecture by NAS while retaining the previously learned knowledge. BP-NAS~\cite{liu2020block}, PolSAR-DNAS~\cite{dong2020automatic} and BI~\cite{xu2019overview} all focus on the architecture designing. BP-NAS proposes a new two-stage NAS method for classic image classification, while PolSAR-DNAS tailors NAS for PolSAR classification task. BI reviews many architecture designing methods about bidirectional intelligence. 
In contrast, the main focus of our work is to introduce NAS into the multi-domain learning task (i.e. adaptively learn what and where to plug adapters). In addition to the current Darts~\cite{liu2018darts} option, our work is quite flexible in using any other NAS alternatives.

To implement a typical NAS algorithm, firstly we need to construct an appropriate operation set, denoted as $O$ containing all possible operations. A graph of architecture containing $M$ nodes are then needed to be defined, where each node is a latent representation (e.g. a feature map in convolutional networks), and each directed edge $(i,j)$ is associated with some operation $o^{(i,j)}$ belonging to $O$.
To make the search space continuous, the categorical choice of a particular operation can be relaxed to the softmax over all possible operations~\cite{liu2018darts}:
\begin{equation}
\label{equ:mix_o}
\overline{o}^{(i,j)}(\cdot) = \sum_{o\in O}^{}\frac{exp(\alpha_{o}^{(i,j)})}{\sum_{o'\in O}^{}exp(\alpha_{o'}^{(i,j)})}o(\cdot),
\end{equation}
where the weights for a pair of nodes are parameterized by a vector $\alpha^{(i,j)}$ of dimension $|O|$, and therefore the architecture searching reduces to learning the variables $\alpha=\{\alpha^{(i,j)}\}$. The variables can be learned with a specific objective function, and the final structure can then be obtained by simply selecting the most likely operation, i.e., $o^{(i,j)}=\arg\max_{o\in O} \alpha_{o}^{(i,j)}$. Then the structure of a model can be denoted by a set of architecture weights.

\begin{table*}[ht]
	\centering
	\caption{Main notations and symbols used throughout the paper.}
	\resizebox{0.77\textwidth}{!}{
		\begin{tabular}{c l l}
			\toprule
			\textbf{Notation} & \multicolumn{2}{c}{\textbf{Definition}} \\
			\midrule
			$D$&  \multicolumn{2}{l}{The number of domains}\\
			
			$\mathcal{D}_d$&  \multicolumn{2}{l}{The $d$-th domain}\\
						
			$(x_d, y_d)$&  \multicolumn{2}{l}{A sample set of class}\\

			$\Psi( \cdot ;\mathcal{A},\mathcal{B}, \Theta)$ &  \multicolumn{2}{l}{The MDL model for all the $D$ domains}\\
			
			$\mathcal{A}$&  \multicolumn{2}{l}{The selected adapter structures (i.e. what to plug) of $\Psi( \cdot ;\mathcal{A},\mathcal{B}, \Theta)$}\\
			
			$\mathcal{B}$&  \multicolumn{2}{l}{the selected adapter plugging strategies (i.e. where to plug) of $\Psi( \cdot ;\mathcal{A},\mathcal{B}, \Theta)$}\\
			
			$\Theta$&  \multicolumn{2}{l}{The parameters of $\Psi( \cdot ;\mathcal{A},\mathcal{B}, \Theta)$}\\
			
			$\Psi_0( \cdot ;\mathcal{A}_0, \mathcal{B}_0, \Theta_0)$&  \multicolumn{2}{l}{The pretrained network as common trunk model for $\Psi( \cdot ;\mathcal{A},\mathcal{B}, \Theta)$}\\
			
			$\Theta_0$&  \multicolumn{2}{l}{The parameters of $\Psi_0( \cdot ;\mathcal{A}_0, \mathcal{B}_0, \Theta_0)$}\\
			
			$N$&  \multicolumn{2}{l}{The number of domain-agnostic layers for $\Psi( \cdot ;\mathcal{A},\mathcal{B}, \Theta)$}\\
			
			$f^n$&  \multicolumn{2}{l}{The $n$-th domain-agnostic layer of $\Psi( \cdot ;\mathcal{A},\mathcal{B}, \Theta)$}\\
			
			$\Psi_d(.;\mathcal{A}_d,\mathcal{B}_d, \Theta_0,\Theta_d^{a})$&  \multicolumn{2}{l}{The adaptation model for $\mathcal{D}_d$}\\
			
			$a_d^n$&  \multicolumn{2}{l}{The adapter to be plugged into the $n$-th location for $\mathcal{D}_d$}\\
			
			$\Theta_d^{a}$&  \multicolumn{2}{l}{The parameters of adapters for $\Psi_d(.;\mathcal{A}_d,\mathcal{B}_d, \Theta_0,\Theta_d^{a})$}\\	
			
			$\mathcal{A}_d$&  \multicolumn{2}{l}{The adapter structure used for $\Psi_d(.;\mathcal{A}_d,\mathcal{B}_d, \Theta_0,\Theta_d^{a})$}\\	
			
			$\mathcal{B}_d$&  \multicolumn{2}{l}{The adapter plugging strategy used for $\Psi_d(.;\mathcal{A}_d,\mathcal{B}_d, \Theta_0,\Theta_d^{a})$}\\	
			
			$\alpha_{d}^n$&  \multicolumn{2}{l}{The structure of $a_d^n$}\\
			\bottomrule		
		\end{tabular}%
	}
	
	\label{Notation}%
\end{table*}%

\section{Method}\label{method}
\subsection{Overview}\label{Problem_Formulation}
To better understand our representations, we provide detailed explanations of the main notations and symbols used throughout this paper as shown in Table~\ref{Notation}.

In MDL, data is sampled from $D$ domains $\{\mathcal{D}_d\}_{d=1}^D$ with corresponding labels for different tasks, and a sample belonging to the $d$-th domain can then be denoted as $(x_d, y_d)$. The goal is to learn a single compact model $\Psi( \cdot ;\mathcal{A},\mathcal{B}, \Theta)$ that works well for all the $D$ domains, which can be addressed in a two-stage fashion shown in Figure~\ref{fig:multi_domain_learning}. Here, $\mathcal{A}$ denotes the selected adapter structures (i.e. what to plug), $\mathcal{B}$ denotes the selected adapter plugging strategies (i.e. where to plug) for all the domains and $\Theta$ represents the parameters. Both $\mathcal{A}$ and $\mathcal{B}$ affect the structure of the learned MDL model.
In the first stage, we choose a pretrained network $\Psi_0( \cdot ;\mathcal{A}_0, \mathcal{B}_0, \Theta_0)$ which consists of $N$ layers as our trunk model:
\begin{equation}
\Psi_0( \cdot ;\mathcal{A}_0, \mathcal{B}_0, \Theta_0) = f^N \circ f^{N-1} \circ \dots \circ f^1( \cdot ;\Theta_0),
\end{equation}
where $\Theta_0$ represents the pretrained parameters and $f^n$ ($n\in\{1,2,\dots,N\}$) denotes the $n$-th domain-agnostic layer. $\mathcal{A}_0$ and $\mathcal{B}_0$ are not utilized because the pretrained network has no adapters. 

In the second stage, for each domain $\mathcal{D}_d$, we need to construct an adaptation model $\Psi_d(.;\mathcal{A}_d,\mathcal{B}_d, \Theta_0,\Theta_d^{a})$ which is composed of the trunk model and a set of additional adapters. We use $\Theta_d^{a}$ to represent the parameters of adapters. The adapter to be plugged into the $n$-th location for domain $\mathcal{D}_d$ is represented as $a_d^n$. For ease of notation, we will apply a domain-agnostic layer followed by its adapter module as $f_d^n = a_d^{n} \circ f_n$. 

After selecting an appropriate adapter structure and obtaining an appropriate plugging strategy for the adaptation model, the goal is then to find an optimal $\Theta_{d}^{a*}$ of the adapters while fixing $\mathcal{A}_d$ and $\mathcal{B}_d$:
\begin{equation}
\label{equ:mdl_train}
\Theta_{d}^{a*} = \mathop{\arg\min}\limits_{\Theta_d^{a}} \sum\limits_{(x_d, y_d) \in \mathcal{D}_d} ||\Psi_d(x_d;\mathcal{A}_d,\mathcal{B}_d, \Theta_0,\Theta_d^{a})-y_d||^2_2,
\end{equation}
where the domain-agnostic parameters $\Theta_0$ is fixed in training and the only term to optimize is $\Theta_d^{a}$, i.e. domain-specific parameters adapted to the domain $\mathcal{D}_d$. At the end, the MDL model $\Psi(\cdot ;\mathcal{A},\mathcal{B}, \Theta)$ is obtained with the fixed domain-agnostic parameters $\Theta_0$ and the optimal domain-specific parameters $\{\Theta_{d}^{a*}\}_{d=1}^D$.

As shown in Figure~\ref{fig:multi_domain_learning} and Equation~\eqref{equ:mdl_train},
the domain-specific adapter $\Theta_d^{a}$ affects the performance primarily among domains, which brings out the extra parameters and complexity. Previous methods usually use the same adapter structure for each domain, while in this work, we claim that an MDL model should be equipped with different adapters that vary from domain to domain.
A simple adapter module would fail in complex domain-transformation, while a complicated one may lead to the waste of computational resources. 
Therefore, it is a challenge to make a balance between model effectiveness and compactness with respect to different domains.

To obtain a discriminative MDL model, we propose to find a set of domain-specific adapter structures for each adaptation model $\Psi_d( \cdot ;\mathcal{A}_d,\mathcal{B}_d, \Theta_0,\Theta_d^{a})$, while taking both the domain differences and complexity into consideration, this will be detailed in Section~\ref{NAS_adaptation}. We also observe that the whole MDL model can be further compacted by removing several specific adapters (i.e. setting those adapters to be an identity mapping) without sacrificing the performance. Our scheme thereby further introduces a selection of plugging strategy to achieve a more compact MDL model in Section~\ref{plugging_strategy}. With our adapter plugging strategy selection, the domain-specific adapter modules are more flexible to different domains. The illustration of our scheme for MDL is shown in Figure~\ref{fig:our_method}.

\subsection{Adapter Module Selection}\label{NAS_adaptation}
In this section, we introduce the process of searching the adapter module structures, taking the domain diversity and the complexity into consideration.

According to Equation \eqref{equ:mdl_train} and Figure~\ref{fig:multi_domain_learning}, finding the specific adapter modules in essence means seeking an appropriate structure for each adapter in $\{a_{d}^n\}^{N}_{n=1}$.
This problem can be reduced to learning a set of structure weights $\{\alpha_{d}^n\}^{N}_{n=1}$ by NAS~\cite{liu2018darts} and then $\mathcal{A}_d $ is correspond to $\{\alpha_d^1,\dots,\alpha_d^N\}$. A NAS-adapter used to select the structure of the MDL adapter $a_{d}^n$ is detailed in what follows.

\begin{figure}[t]
	\centering
	\includegraphics[width=0.49\textwidth]{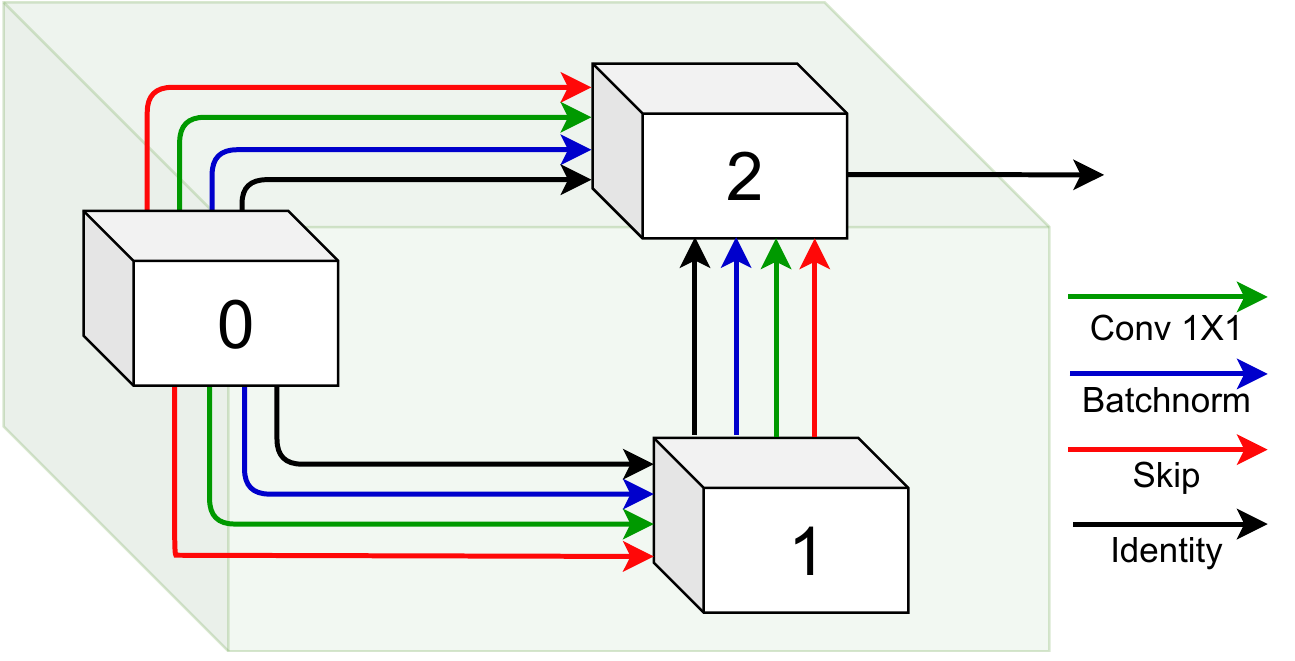}
	\caption{Illustration of our NAS-adapter. A NAS-adapter consists of $M$ nodes ($M=3$ in this figure). The final structure is achieved by optimizing the structure parameters and selecting an operation from a possible set. We, in this paper, construct the set with $1\times1$ convolution operation, batch normalization operation, skip connection, and identity shortcut.}
	\label{fig:NAS_adapter}
\end{figure}

To achieve a compact MDL model, the searching space needs to be properly designed, because the additional complexity and performance of the adaptation model $\Psi_d(\cdot;\mathcal{A}_d, \mathcal{B}_d,\Theta_0,\Theta_d^{a})$ depend on the adapter structures $\{a_d^n\}^{N}_{n=1}$.
The core of multi-domain learning (MDL) is to ensure the simplicity and compactness of architecture for the concern of practicability.
To achieve that, we adopt NAS with parameter-constrained prior which limits the searching space of the adapter structure (i.e. what to plug).
We draw inspiration from previous popular structures of adapters~\cite{rebuffi2017learning,rebuffi2018efficient} and collect the operation set $O_a$: $1\times1$ convolution operation, batch normalization operation, skip connection and identity shortcut~\cite{he2016identity}. Our NAS-adapter consists of $M$ nodes, the example for $M=3$ is illustrated in Figure~\ref{fig:NAS_adapter}.

For the domain $\mathcal{D}_d$, we search an appropriate set of adapter structures by learning a set of structure weights $\mathcal{A}_d$:
\begin{equation}
\label{equ:alpha}
\mathcal{A}_{d}^{*} = \mathop{\arg\min}\limits_{\mathcal{A}_d} \sum\limits_{(x_d, y_d) \in \mathcal{D}_d} ||\Psi_d(x_d;\mathcal{A}_d,\mathcal{B}_d,\Theta_0,\Theta_d^{a})-y_d||^2_2.
\end{equation}

We utilize a similar training strategy in~\cite{liu2018darts} and the training data is equally divided into a validation set and a training set. Specifically, we use the validation set to update the weights of adapter structures, while the parameters of adapters are optimized by the training set. 
\begin{table}[b]
		\centering
	\caption{Performance of different adapter structures with the trunk model VGG-16.}
\resizebox{0.5\textwidth}{!}{
	\begin{tabular}{lccccc}
		\toprule
		Method  &Flowers&FGVC &CIFAR100 &MITIndoor&Total Param.\\			
		\midrule
		BN~\cite{bilen2017universal}&91.47\%&63.04\%&64.80\%&57.60\%&\textbf{$\approx$1}\\
		DAN~\cite{rosenfeld2018incremental}&\textbf{92.65\%}&\textbf{86.80\%}&\textbf{74.45\%}&\textbf{63.02\%}&2.05\\
		\bottomrule						
	\end{tabular}
}
	\label{tab:adapt_structure}
\end{table}	

Through searching the adapter structure for different domains, our multi-domain learning method aims to build an effective multi-domain model with limited memory cost, that is, keeping a good balance between the effectiveness and efficiency. More specifically, our method is able to automatically determine the adapter structure according to the complexity of domain. Namely, it seeks for selecting simple adapters for simple domains to save memory cost while selecting complicated adapters for complex domains to pursue the performance. Typically, a complex domain corresponds to the dataset that comprises more content-diverse samples with richer textures and complicated background clutters, following a very large multi-class classification problem setting with massive samples. In contrast, conventional MDL methods usually adopt the same predefined handcrafted adapter structure for different domains (e.g. BN~\cite{bilen2017universal}, DAN~\cite{rosenfeld2018incremental}). As a result, they can either build a complicated adapter structure with high accuracy to deal with complex domains or only construct a simple adapter structure but with low accuracy for all domains. Thus, they are incapable of achieving a good trade-off between effectiveness and efficiency. Suppose a dataset consists of both complex domains (e.g. MITIndoor) and simple domains (e.g. Flowers).
In this situation, adopting the same predefined handcrafted adapter structure for different domains and learning the parameters only is not enough. As shown in Table~\ref{tab:adapt_structure}, adopting a simple adapter BN and learning the parameters of it can achieve high performance on a simple domain (i.e. Flowers) but low performance on complex domains (i.e. FGVC, CIFAR100, MITIndoor). And adopting a complicated adapter DAN and learning the parameters of it can achieve high performance on all domains but consumes more memory cost. In contrast, our method can adaptively build adapter structures for different domains and is able to keep a good trade-off between the effectiveness and efficiency in such a situation.

\subsection{Plugging Strategy Selection}\label{plugging_strategy}
The learning requirements vary for different domains. Therefore, using the same plugging strategy may not be optimal for all domains. Prior works plug the adapters into every available slot of MDL models, leaving much room for the improvement of compactness, achieving that is the focus of this section. Given a set of adapters with fixed searched structures, we further propose to determine whether or not to plug the adapter into each possible slot for every visual domain with NAS. 

This problem can be considered as a typical NAS problem where the operation set only consists of the identity operation and the candidate adapter operation. Therefore, for the domain $\mathcal{D}_d$, searching a plugging strategy is reduced to learning a set of continuous variables $\mathcal{B}_d=\{\beta_d^1,\beta_d^2,...,\beta_d^N\}$ for the $N$ possible plugging locations:
\begin{equation}
\label{equ:beta}
\mathcal{B}_d^* = \mathop{\arg\min}\limits_{\mathcal{B}_d} \sum\limits_{(x_d, y_d) \in \mathcal{D}_d} ||\Psi_d(x_d;\mathcal{A}_d,\mathcal{B}_d,\Theta_0,\Theta_d^{a})-y_d||^2_2,
\end{equation}

In the training process for domain $\mathcal{D}_d$, we need to optimize the weights of the plugging strategy $\mathcal{B}_d = \{\beta_d^1,\dots,\beta_d^N\}$ and the parameters of added adapters $\Theta_d^a$, which is a bilevel optimization problem. Similarly to the optimization done in Section~\ref{NAS_adaptation}, we use the validation set to update the weights of the plugging strategy, and the training set for the adapter parameters. 

After learning an appropriate adapter structure and an appropriate adapter plugging strategy, we can further update the parameters $\Theta_d^{a}$ with the training data by Equation \eqref{equ:mdl_train}.

We have provided an algorithm flow (i.e. Algorithm~\ref{nas_ada}) and training details about our method. 
	Our method mainly consists of two steps: adapter module selection (line 2-7) and plugging strategy selection (line 8-13). 
	For the adapter module selection, we first construct the training set, validation set, and adapter module structure searching space (line 2-3), then learning the adapter module structure and parameters (line 4-7). For the plugging strategy selection, we first construct the plugging strategy searching space (line 8), then learning the plugging strategy and adapter parameters (line 9-12), finally we fix the adapter structures and plugging strategy and continue to update the parameters of adapters with the training data (line 13).

	\begin{algorithm}[ht]
	\caption{Our NAS-driven MDL method}
	\label{nas_ada}
	\KwIn{The data for $D$ domains $\{\mathcal{D}_d\}_{d=1}^D$, a pretrained network $\Psi_0( \cdot ;\mathcal{A}_0, \mathcal{B}_0, \Theta_0)$ which consists of $N$ domain-agnostic layers $\{f_n\}_{n=1}^{N}$ as the trunk model and the maximum number $T_{max}$ of iterations.}
	\everypar={\nl}
	\For{domains $1, 2, 3, \dots, D$}{
		Divide the training data (the sample $(x_d, y_d)$ is from domain $\mathcal{D}_d$) into a training set and validation set equally\;
		\tcp{Adapter Module Selection}
		\everypar={\nl}
		Create nodes and corresponding edges of the NAS-adapter parametrized by $\alpha_{d}^n$ after each domain-agnostic layer $f_n$\;
		\For{iterations $1, 2, 3, \dots, T_{max}$}{
			Update the structure weights $\mathcal{A}_d=\{\alpha_{d}^n\}_{n=1}^N$ with data sampled from the validation set by Equation(4)\;
			Update the parameters $\Theta_d^{a}$ with data sampled from the training set by Equation (3)\;	
		}
		\tcp{Plugging Strategy Selection}
		\everypar={\nl}
		Create the operations parametrized by $\beta_d^n$ after each domain-agnostic layer $f_n$\;
		\For{iterations $1, 2, 3, \dots, T_{max}$}{
			Update the weights of the plugging strategy $\mathcal{B}_d=\{\beta_d^n\}_{n=1}^{N}$ with data sampled from the validation set by Equation (5)\;
			Update the parameters $\Theta_d^a$ with data sampled from the training set by Equation (3)\;
		}
		Fix the adapter structure and plugging strategy, update the parameters $\Theta_d^a$ with data sampled from the domain $\mathcal{D}_d$ by Equation (3)\;
	}
	\KwOut{Derive the MDL model $\Psi(\cdot ;\mathcal{A},\mathcal{B}, \Theta)$ with the fixed domain-agnostic parameters $\Theta_0$ and the optimal domain-specific parameters $\{\Theta_{d}^{a*}\}_{d=1}^D$, a set of adapter structure weights $\{\mathcal{A}_d^*\}_{d=1}^D$ and plugging strategy$\{\mathcal{B}_d^*\}_{d=1}^D$.}
\end{algorithm}

\section{Experiments}\label{experiments} 
\subsection{Datasets}
We evaluated our approach with two different benchmarks. We first use the Visual Decathlon benchmark~\cite{rebuffi2017learning}, built with $10$ different datasets from \textbf{ImageNet}~\cite{russakovsky2015imagenet} to \textbf{German Traffic Signs}~\cite{stallkamp2012man}, in which the images are resized to $72 \times 72$.
As for the second benchmark~\cite{li2019efficient}, a set of seven popular vision datasets are collected for evaluation and this benchmark is used for large CNNs. \textbf{SUN 397}~\cite{xiao2010sun} contains $397$ classes of scene images and more than a million images. \textbf{MITIndoor}~\cite{valenti2007indoor} is an indoor scene dataset with $67$ classes and $80$ samples per class. \textbf{FGVC-Aircraft Benchmark}~\cite{bilen2017universal} is a fine-grained classification dataset of $10,000$ images of $100$ types of airplanes. \textbf{Flowers102}~\cite{nilsback2008automated} is a fine-grained dataset with $102$ flower categories and $40$ to $258$ images per class. \textbf{CIFAR100}~\cite{krizhevsky2009learning} contains $60,000$ tiny images, from 100 classes. \textbf{Caltech256}~\cite{griffin2007caltech} contains $30,607$ images of $256$ object categories, with at least $80$ samples per class. \textbf{SVHN}~\cite{netzer2011reading} is a digit recognition dataset with $10$ classes and more than $70,000$ samples. In this benchmark, images are rescaled to a common size of $224\times224$ and the training and testing sets are defined by the corresponding dataset, if available, while $75\%$ of samples are used for training and $25\%$ are for testing, otherwise.

\subsection{Implementation Details}
\paragraph{Network architectures} 
For the Visual Decathlon benchmark, we follow~\cite{rebuffi2018efficient} and conduct experiments using a ResNet~\cite{he2016identity} with $26$ layers as the common trunk structure. We employ the same data pre-processing setting and freeze the parameters of our \textbf{ResNet-26} model after the pretraining on ImageNet. For the second benchmark, we follow~\cite{li2019efficient} and use a \textbf{VGG-16}~\cite{simonyan2014very} model in all experiments. This model contains convolutional layers of dimension ranging from $64$ to $4096$, and the parameters are also pretrained on ImageNet.

\paragraph{Evaluation protocol} These two benchmarks are designed to address classification problems. Similar to~\cite{rebuffi2017learning,rebuffi2018efficient,li2019efficient}, we report the accuracy for each domain (denoted by ``Acc.") and the average accuracy over all the domains (denoted by ``Ave. Acc.") . The score function~\cite{rebuffi2017learning} $S$ (denoted by ``S.") is also adopted for the evaluation, formulated as:
\begin{equation}
S=\sum_{d=1}^{N}\lambda_d\max\{0,E^{max}_d-E_d\}^2,
\end{equation}
where $N$ is the number of different domains and $E_d$ denotes the test error of the MDL method for the domain $D_d$. $E^{max}_d$ is twice the testing error rate of baseline, which is the fully finetuned network for that domain, and $\lambda_d$ is a coefficient to ensure the best result for each domain is $1000$. The score favors the methods that perform well over all domains, and methods which are outstanding only on a few domains will be penalized. Furthermore, parameter cost is also taken into consideration following~\cite{li2019efficient, rebuffi2017learning}. 
We report the adapter parameter usage for each domain (denoted by ``Ada. Param.") or
report the total number of parameters relative to the initial pretrained trunk model (excluding the classifiers) over all domains (denoted by ``Total Param.").

\paragraph{Training details}
For the Visual Decathlon benchmark, we train the ResNet-26 model with the same training strategy in~\cite{rebuffi2017learning}. To select a NAS-adapter module structure, we follow the strategy in~\cite{liu2018darts} with an NVIDIA 1080Ti GPU and divide the training dataset into two parts of equal size. One part is used to optimize the structure weights while the other part is to optimize the network parameters. For structure weights learning, we use Adam optimizer~\cite{kingma2014adam} with weight decay $0.001$ and momentum ($0.5$, $0.999$) and the initial learning rate set to $0.0003$.
For network parameters optimization, we use SGD optimizer with an initial learning rate $0.01$ (annealed down to zero following a cosine schedule without restart~\cite{loshchilov2016sgdr}), momentum $0.9$, and weight decay $0.0005$.
For plugging strategy selection, the learning rate is initialized by $0.005$ and divided by $10$ after $20$, $40$, $60$ epochs. For the second benchmark, we utilize the training approach in~\cite{li2019efficient} for the VGG-16 model. The rest of the settings for adapter module selection and plugging strategy selection are the same as those on the former.
The elapsed time taken for running our NAS-driven MDL method mainly consists of three parts: the NAS training time for adapter structure selection (denoted by ``NAS-adapter Time"), the NAS training time for plugging strategy selection (denoted by ``NAS-plugging Time"), and the training time for adapter parameters updating (denoted by ``Adapter-parameters Time"). On the Visual Decathlon benchmark with the trunk model ResNet-26, the total elapsed time for running our method is about 36 hours (NAS-adapter Time 16, NAS-plugging Time 8, Adapter-parameters Time 12). On the benchmark of seven domains with the trunk model VGG-16, the total elapsed time for running our method is about 185 hours (NAS-adapter Time 80, NAS-plugging Time 40, Adapter-parameters Time 65) with an NVIDIA 1080Ti GPU.

\subsection{Ablation Study}
In this section, we firstly carry out ablation experiments to validate the effectiveness of our proposed NAS-adapter module (\textbf{ablation experiment-1}) and plugging strategy selection scheme (\textbf{ablation experiment-2}). Then we give a detailed analysis about what to plug (i.e. the adapter structure) and where to plug (i.e. the plugging strategy). About what adapter structure to plug, we give a statistic for the distribution of learned adapter structure across domains (\textbf{ablation experiment-5}) and show the importance of adapter structure search (\textbf{ablation experiment-3}). About where the adapters to plug, we give a statistic for the frequency of each plugging location to be selected (\textbf{ablation experiment-6}) and compare our selected plugging strategy with other hand-crafted plugging strategies (\textbf{ablation experiment-4}). Finally, we make a discussion about the accuracy of our method with regard to the domain diversity (\textbf{ablation experiment-7}) and different paradigms for the task of multi-domain learning (\textbf{ablation experiment-8}).


\begin{table}[!t]
	\centering
	\caption{Accuracy of different adapter modules with trunk model VGG-16, using ``All" plugging strategy.}
	\resizebox{0.5\textwidth}{!}{
		\begin{tabular}{lcccc}
			\toprule
			Adapter Structure&Plugging Strategy  &MITIndoor &Flowers& FGVC\\
			\midrule
			Res Adapt  &\multirow{4}{*}{All}&72.40$\pm$0.24\% & 96.43$\pm$0.12\%& 88.92$\pm$0.28\%\\
			$1 \times 1$ Adapt&~  &63.02$\pm$ 0.26\% &92.65$\pm$0.16\%& 86.80$\pm$0.32\% \\	
			BN Adapt&~  &57.60$\pm$0.15\% &91.47$\pm$0.11\% & 63.04$\pm$0.25\%\\		
			NAS Adapt&~  &\textbf{73.05$\pm$ 0.26\%} &\textbf{96.81$\pm$0.15\%}& \textbf{89.08$\pm$0.33\%}\\
			\bottomrule
		\end{tabular}
	}
	
	\label{tab:one}
\end{table}

\begin{table}[!t]
	\centering
	\caption{Accuracy of different adapter modules with trunk model ResNet-26, using ``All" plugging strategy.}	
	\resizebox{0.5\textwidth}{!}{
		\begin{tabular}{lccccc}
			\toprule
			Adapter Structure &Plugging Strategy& OGlt &SVHN&DTD \\
			\midrule
			Res Adapt&\multirow{4}{*}{All}  & 89.82$\pm$0.13\% &96.17$\pm$0.09\%&57.02$\pm$0.19\% \\
			$1 \times 1$ Adapt&~ & 89.67$\pm$0.16\% &96.77$\pm$0.12\% &56.54$\pm$0.22\%\\	
			BN Adapt&~  & 84.83$\pm$0.11\% & 94.10$\pm$0.07\%&51.60$\pm$0.13\%\\		
			NAS Adapt&~  & \textbf{90.02$\pm$0.15\%} &\textbf{96.98$\pm$0.11\%}&\textbf{59.30$\pm$0.20\%}\\
			\bottomrule
		\end{tabular}
	}
	
	\label{tab:two}
\end{table}

Previous MDL approaches construct the adaptation module with a fixed hand-crafted structure and directly add an adaptation module after each layer of common trunk model. We select three different adapter structures (Res Adapt~\cite{rebuffi2017learning}, $1 \times 1$ Adapt~\cite{rosenfeld2018incremental}, BN Adapt~\cite{bilen2017universal}) and plugging the adapters at each possible slot (such a plugging strategy denoted by ``All") as baselines.

\begin{table}[t]
	\centering
	\caption{Accuracy and adapter parameter usage (setting plugging all 15 adapters to be 100\%) of different adapter structures with trunk of VGG-16. The highest accuracy is in \textbf{bold}, and the lowest parameter usage is \underline{underlined}.}
	\resizebox{1\columnwidth}{!}{
		\begin{tabular}{lccccc}
			\toprule
			Adapter Structure &Plugging Strategy & Evaluation   &MITIndoor &Flowers& FGVC\\
			\midrule
			\midrule
			\multirow{4}{*}{Res Adapt}&\multirow{2}{*}{All}  & Acc.&72.40$\pm$0.24\%& 96.43$\pm$0.12\%& 88.92$\pm$0.30\%\\
			~&~&Ada. Param.&35.42M (100\%) & 35.42M (100)\% & 35.42M (100\%) \\
			\cline{2-6}
			~&\multirow{2}{*}{Ours} & Acc.&\textbf{72.61$\pm$0.26\%}&\textbf{96.66$\pm$0.13\%}&\textbf{88.93$\pm$0.30\%}\\
			~&~&Ada. Param.&\underline{17.32M (48.91\%)} &\underline{16.80M (47.42\%)}&\underline{18.29M (51.65\%)}  \\
			\midrule
			\multirow{4}{*}{$1 \times 1$ Adapt}&\multirow{2}{*}{All}  & Acc. &63.02$\pm$0.26\% &92.65$\pm$0.16\% & 86.80$\pm$0.32\%\\	
			~&~& Ada. Param. & 35.42M (100\%) &35.42M (100\%) &35.42M (100\%)\\	
			\cline{2-6}
			~&\multirow{2}{*}{Ours}  & Acc. &\textbf{67.96$\pm$0.27\%} &\textbf{94.83$\pm$0.18\%}& \textbf{87.42$\pm$0.35\%}\\	
			~&~& Ada. Param. &\underline{7.61M (21.49\%)} &\underline{16.80M (47.43\%)}&\underline{10.74M (30.3\%)}  \\			
			\midrule
			\multirow{4}{*}{BN Adapt} &\multirow{2}{*}{All} & Acc. &57.60$\pm$0.15\% &91.47$\pm$0.11\%&63.04$\pm$0.25\%\\
			~&~& Ada. Param.&60 (100\%) &60 (100\%) &60 (100\%)\\
			\cline{2-6}	
			~&\multirow{2}{*}{Ours} & Acc. &\textbf{68.71$\pm$0.16\%} & \textbf{92.75$\pm$0.12\%}& \textbf{68.29$\pm$0.28\%}\\	
			~&~&Ada. Param. &\underline{24 (40.00\%)} &\underline{15 (25.00\%)}&\underline{26 (43.33\%)}\\					
			\bottomrule
		\end{tabular}
	}
	\label{tab:three}
\end{table}

\begin{table}[t]
	\centering
	\caption{Accuracy and adapter parameter usage (setting plugging all 25 adapters to be 100\%) of different adapter structures with trunk of ResNet-26. The highest accuracy is in \textbf{bold}, and the lowest parameter usage is \underline{underlined}.}
	\resizebox{1\columnwidth}{!}{
		\begin{tabular}{lccccc}
			\toprule
			Adapter Structure &Plugging Strategy & Evaluation  & OGlt  &CIFAR100&DTD\\
			\midrule
			\midrule
			\multirow{4}{*}{Res Adapt}&\multirow{2}{*}{All}  & Acc.  &89.82$\pm$0.13\% &81.31$\pm$0.09\%&57.02$\pm$0.18\%\\
			~&~&Ada. Param.&0.69M (100\%) &0.69M (100\%) & 0.69M (100\%) \\
			\cline{2-6}
			~&\multirow{2}{*}{Ours} & Acc. & \textbf{89.96$\pm$0.17\%} &\textbf{81.45$\pm$0.12\%}&\textbf{57.93$\pm$0.25\%}\\
			~&~&Ada. Param.&\underline{0.49M (70.79\%)}  &\underline{0.30M (42.98\%)}&\underline{0.28M (40.59\%)} \\
			\midrule
			\multirow{4}{*}{$1 \times 1$ Adapt}&\multirow{2}{*}{All}  & Acc. & \textbf{89.67$\pm$0.16\%} &\textbf{80.07$\pm$0.10\%} &56.54 $\pm$ 0.22 \%\\	
			~&~&Ada. Param.&0.69M (100\%) &0.69M (100\%) & 0.69M (100\%) \\
			\cline{2-6}
			~&\multirow{2}{*}{Ours} & Acc.  & 89.39$\pm$0.14\% &79.44$\pm$0.08\%&\textbf{56.98$\pm$0.20\%}\\	
			~&~&Ada. Param.&\underline{0.62M (89.90\%}  &\underline{0.34M (49.93\%)}&\underline{0.28M (39.97\%)}\\
			\midrule
			\multirow{4}{*}{BN Adapt} &\multirow{2}{*}{All} & Acc.  & \textbf{84.83$\pm$0.11\%} & 78.62$\pm$0.06\%&\textbf{51.60$\pm$0.13\%}\\
			~&~&Ada. Param.&100 (100\%) &100 (100\%) &100 (100\%) \\	
			\cline{2-6}
			~&\multirow{2}{*}{Ours}& Acc.  & 83.90$\pm$0.12\% &\textbf{78.75$\pm$0.05\%}&51.54$\pm$0.12\%\\	
			~&~&Ada. Param.& \underline{48 (48.00\%)} &\underline{61 (61.00\%)}&\underline{63 (63.00)\%} \\								
			\bottomrule
		\end{tabular}
	}
	\label{tab:four}
\end{table}

\begin{figure*}[t]
	\centering
	\subfigure{}{
		\begin{minipage}[ht]{0.32\textwidth}
			\includegraphics[width = 1\columnwidth]{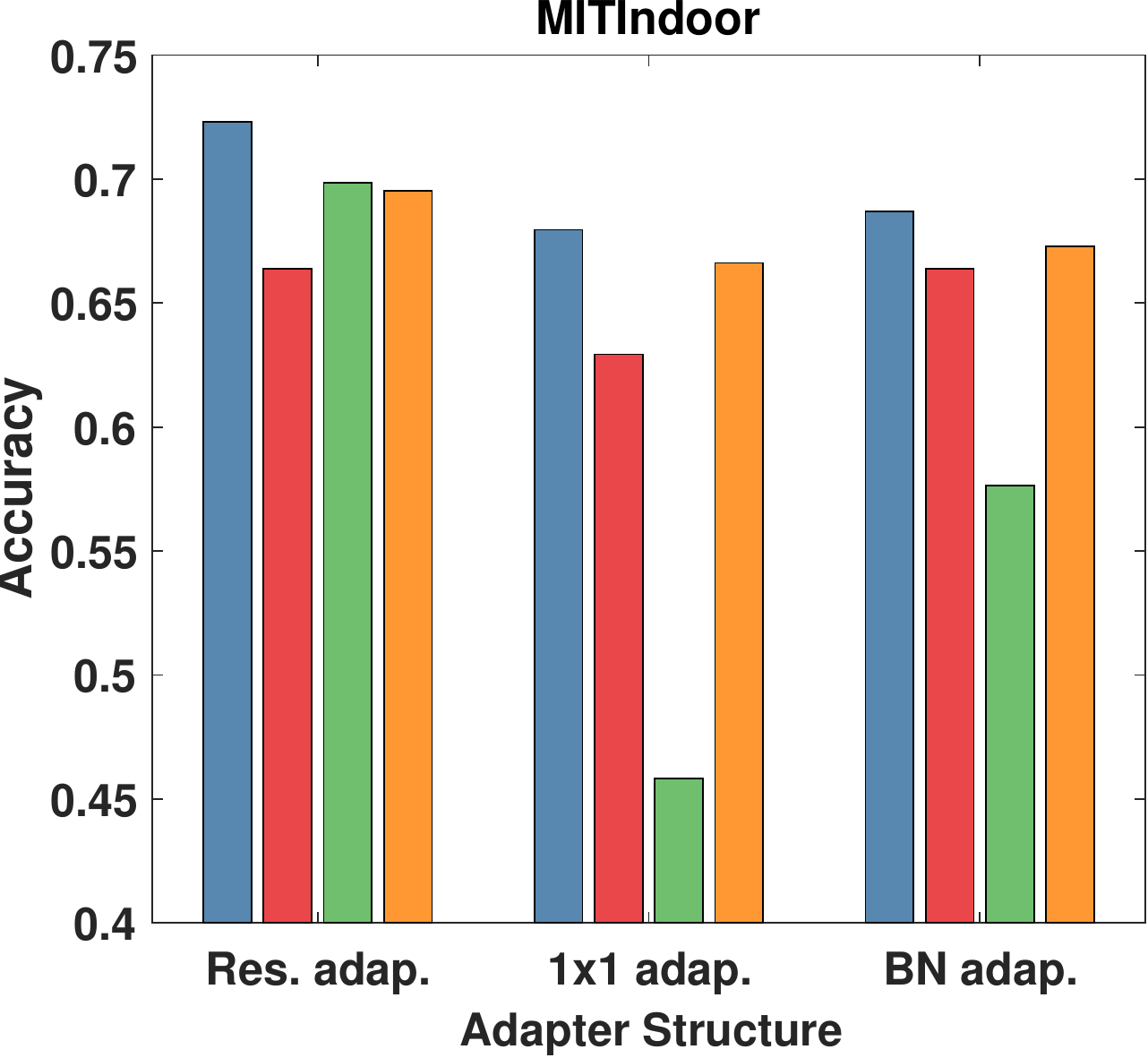}	
	\end{minipage}}
	\subfigure{}{
		\begin{minipage}[ht]{0.32\textwidth}
			\includegraphics[width = 1\columnwidth]{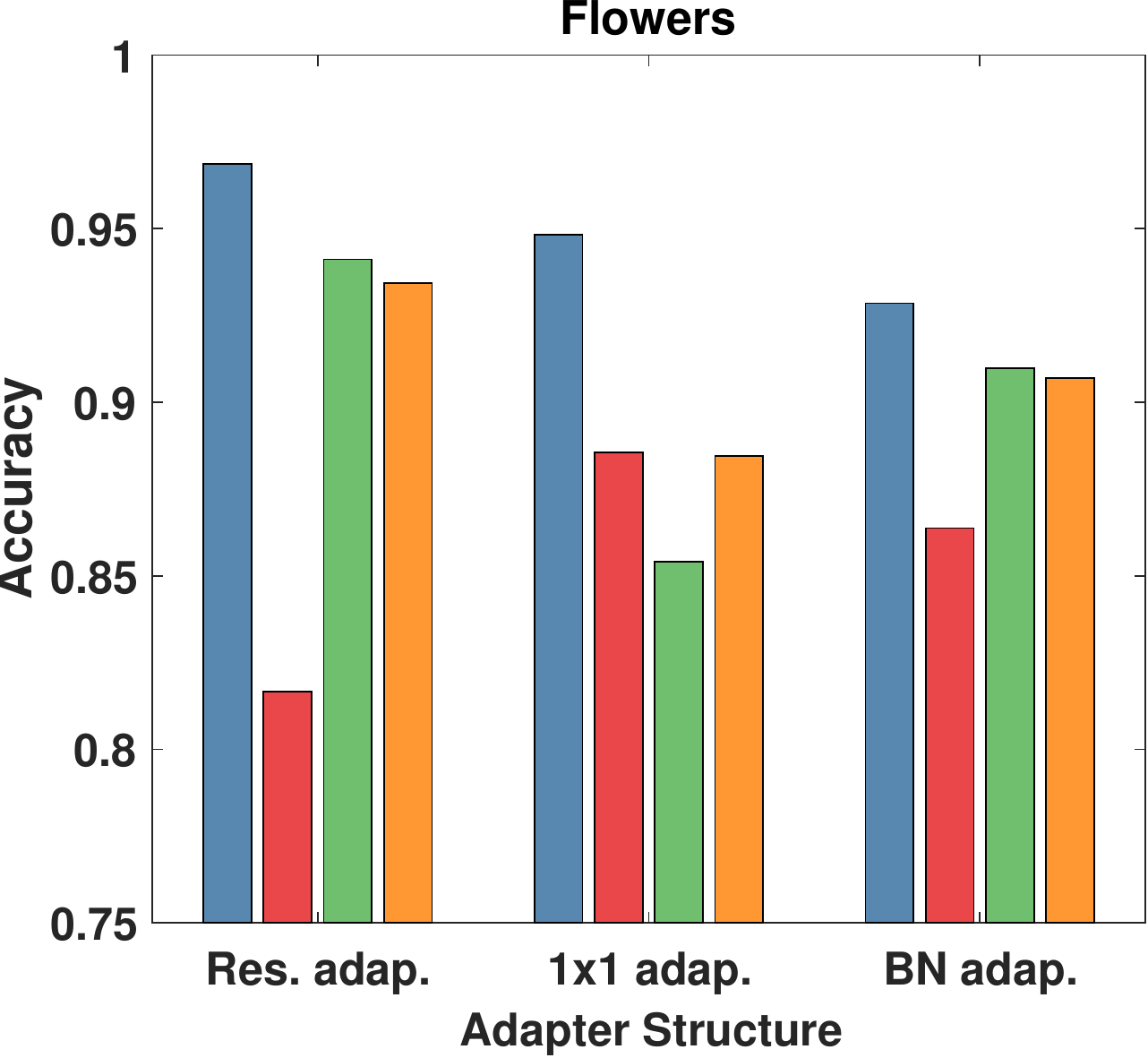}
		\end{minipage}	
	}
	\subfigure{}{
		\begin{minipage}[ht]{0.32\textwidth}
			\includegraphics[width = 1\columnwidth]{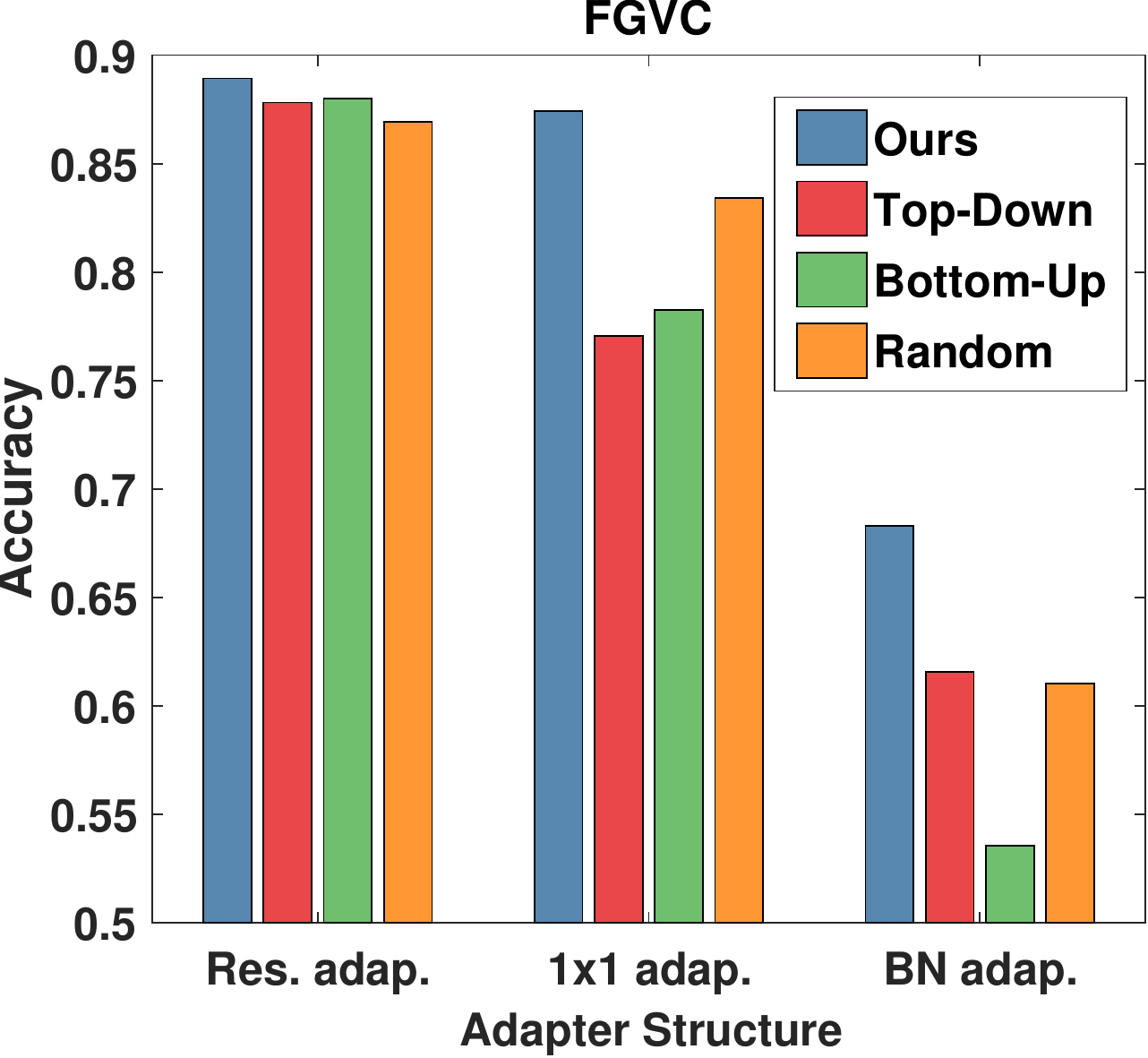}	
		\end{minipage}
	}
	\caption{Performance of different plugging strategies (with the same number of adapter modules) on different datasets (VGG-16). \emph{Ours:} the selected plugging strategy. \emph{Top-Down:} select the first $n$ locations to plug in. \emph{Bottom-Up:} select the last $n$ locations to plug in. \emph{Random:} Randomly select $n$ locations to plug in.}
	\label{fig:four}
\end{figure*}

\begin{figure*}[t]
	\centering
	\subfigure{}{
		\begin{minipage}[ht]{0.32\textwidth}
			\includegraphics[width = 1\columnwidth]{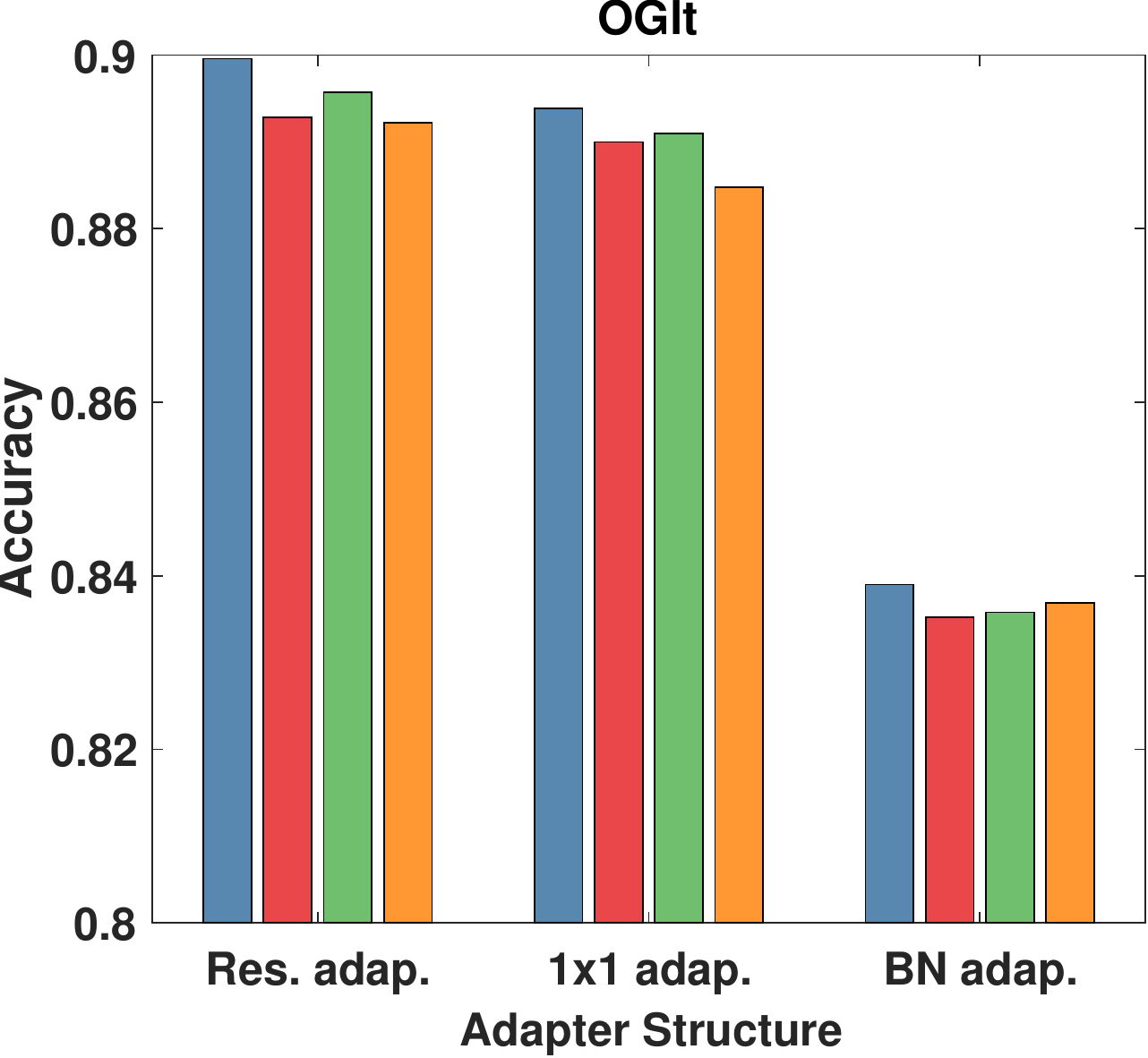}	
	\end{minipage}}
	\subfigure{}{
		\begin{minipage}[ht]{0.32\textwidth}
			\includegraphics[width = 1\columnwidth]{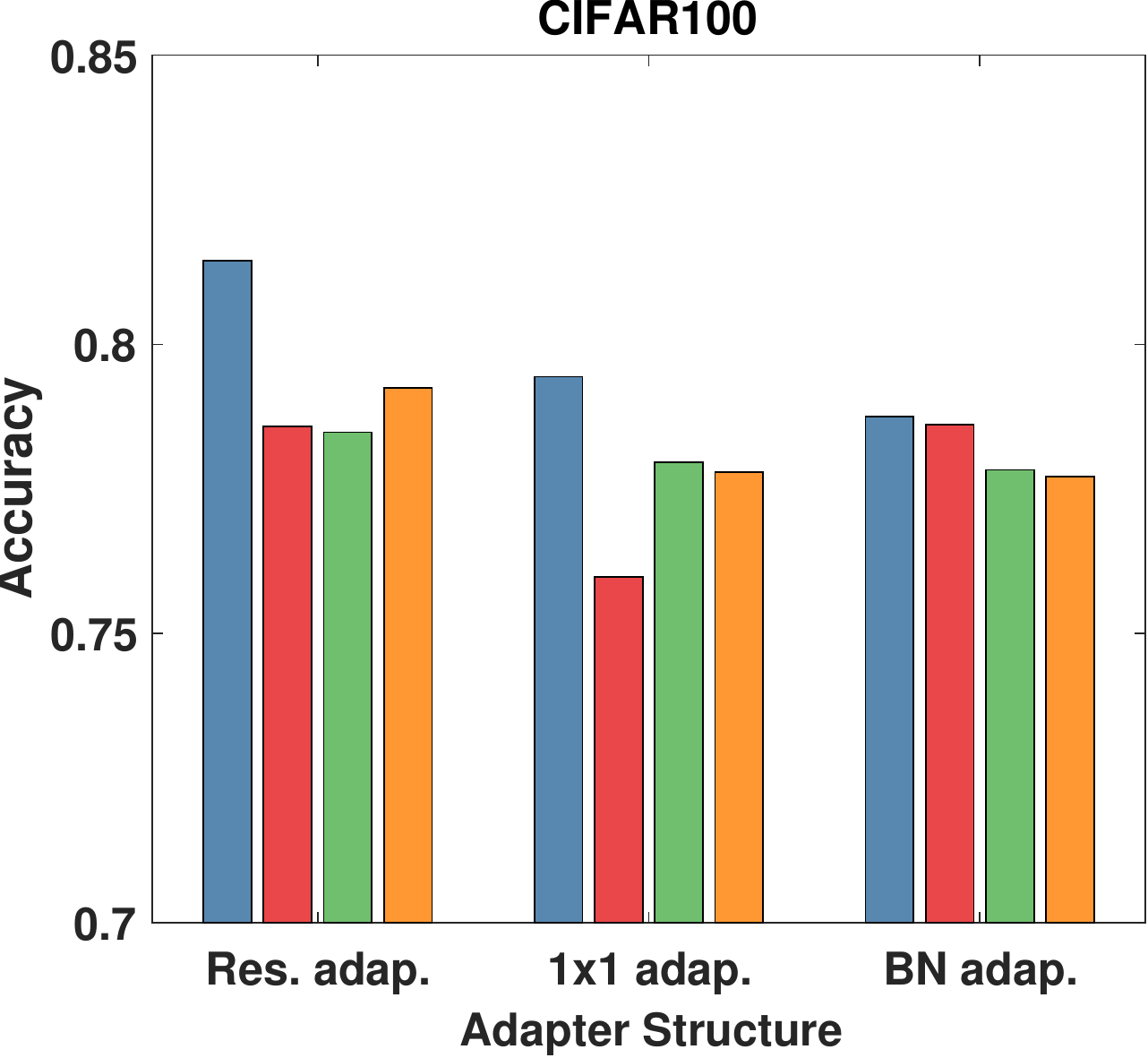}
		\end{minipage}	
	}
	\subfigure{}{
		\begin{minipage}[ht]{0.32\textwidth}
			\includegraphics[width = 1\columnwidth]{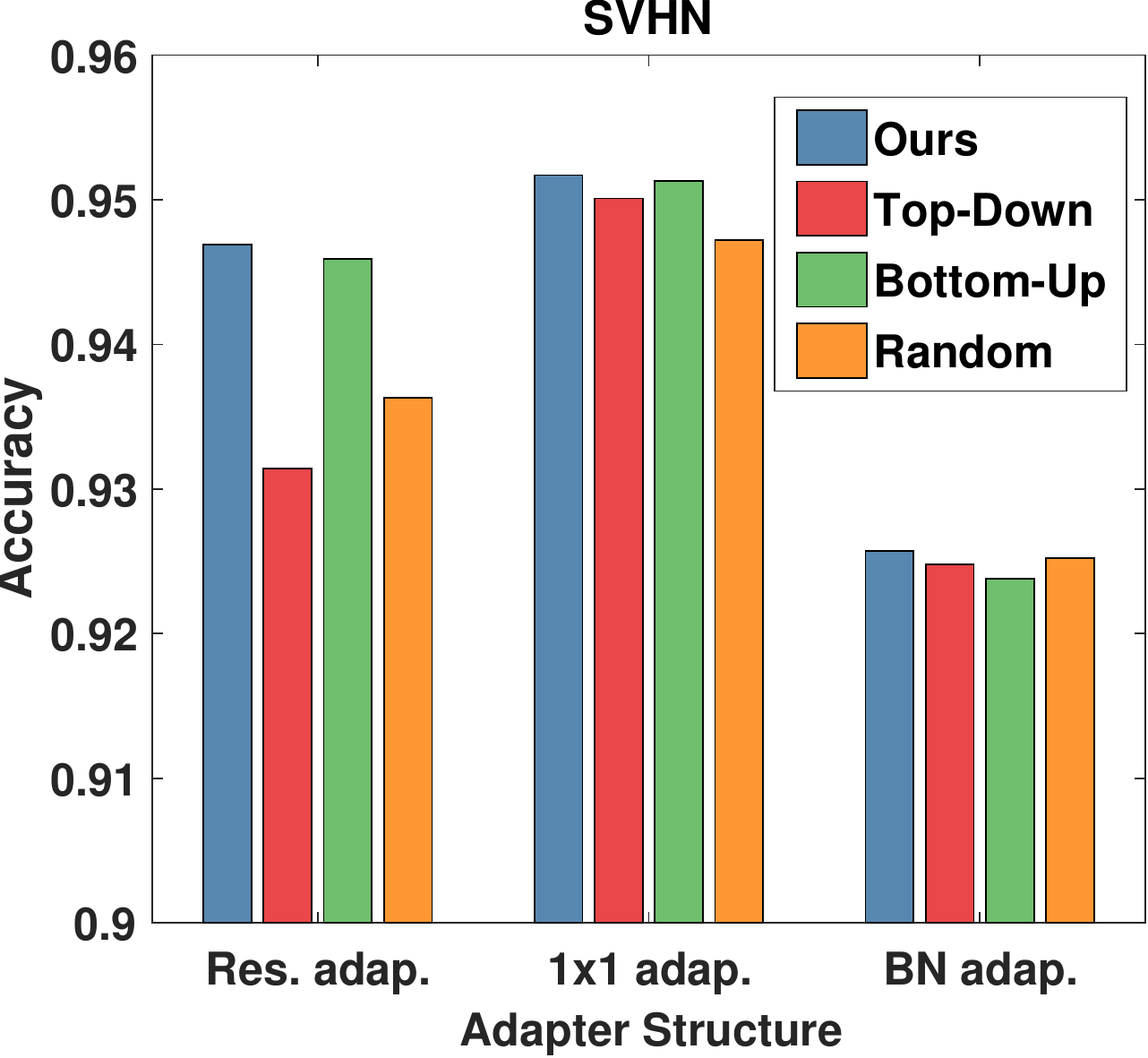}	
		\end{minipage}
	}
	\caption{Performance of different plugging strategies (with the same number of adapter modules) on different datasets (ResNet-26). \emph{Ours:} the selected plugging strategy. \emph{Top-Down:} select the first $n$ locations to plug in. \emph{Bottom-Up:} select the last $n$ locations to plug in. \emph{Random:} randomly select $n$ locations to plug in.}
	\label{fig:five}
\end{figure*}

\begin{table}[t]
	\centering
	\caption{Performance of our method with or without adapter structure search for VGG-16.}
	\resizebox{1\columnwidth}{!}{
		\begin{tabular}{lcccccc}
			\toprule
			Plugging Strategy & Adapter Structure Search &Adapter Structure  &MITIndoor &Flowers& FGVC & Total Param.\\
			\midrule
			\multirow{4}{*}{Ours}	&\multirow{3}{*}{no}&Res Adapt&72.61\%&96.66\%&88.93\%&1.79\\
			&&$1 \times 1$ Adapt&67.96\%&94.83\%&87.42\%&1.79\\
			&&BN Adapt&68.71\%&92.75\%&68.29\%&\textbf{$\approx$1}\\
			\cline{2-7}
			&yes&NAS Adapt&\textbf{73.51\%}&\textbf{96.96\%}&\textbf{89.34\%}&1.33\\
			\bottomrule
		\end{tabular}
	}
	\label{tab:vgg:plugging}
\end{table}		

\subsubsection{Comparison with hand-crafted adapter structures} We compare our NAS-adapter module with other three hand-crafted ones in Table~\ref{tab:one} and Table \ref{tab:two}. For a fair comparison, we also construct the adaptation model by embedding the NAS-adapter module after each domain-agnostic layer (i.e. NAS Adapt). While taking VGG-16 as the common trunk model in Table~\ref{tab:one}, we can observe that the adaptation model with our NAS-adapter can achieve the best results among others. Res Adapt yields the second place of those three datasets, followed by $1 \times 1$ Adapt and BN Adapt. As for experiments where ResNet-26 serves as the common trunk model in Table~\ref{tab:two}, our NAS-adapter module still performs better than others.
\begin{figure}[!t]
	\centering
	\includegraphics[width=0.5\textwidth]{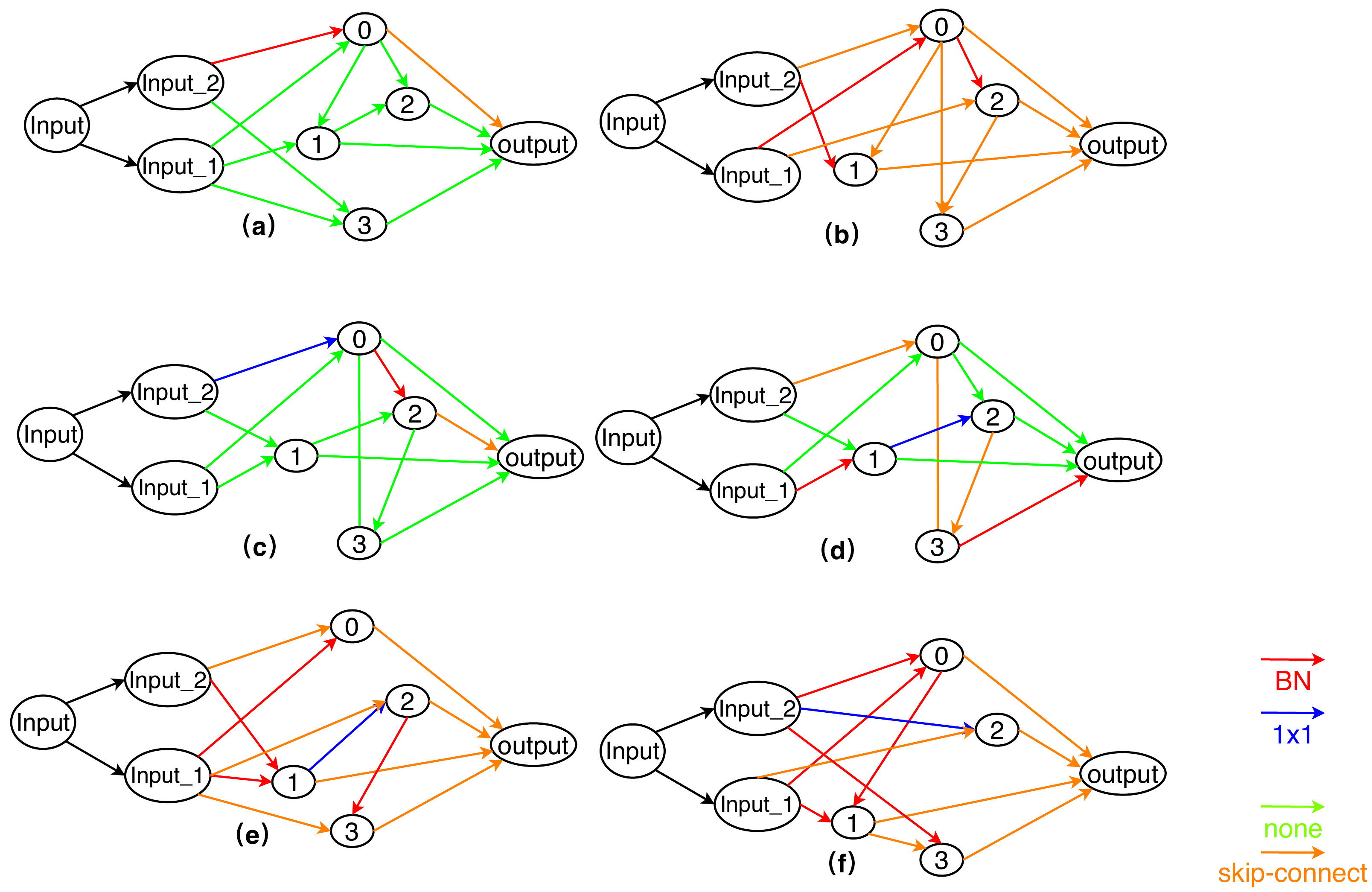}
	\caption{(a): BN Adapt~\cite{bilen2017universal}, (b): NAS-1, (c): $1 \times 1$ Adapt~\cite{rosenfeld2018incremental}, (d): Res Adapt~\cite{rebuffi2017learning}, (e): NAS-2, (f): NAS-3.}
	\label{rfig:novel_module}
\end{figure}

\begin{figure*}[t]
	\centering
	\subfigure{}{
		\begin{minipage}[t]{1\textwidth}
			\includegraphics[width = 1\columnwidth]{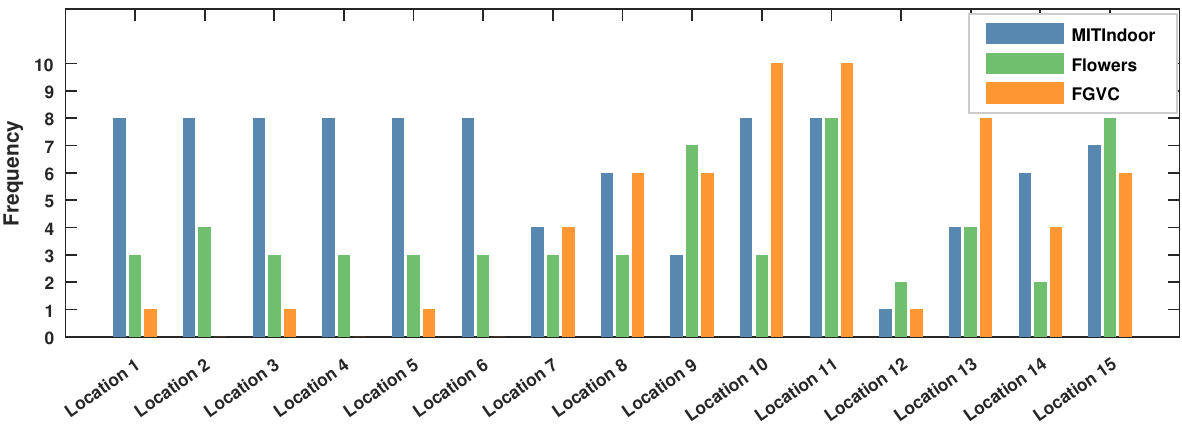}	
	\end{minipage}}
	\caption{Frequency of each plugging location to be selected on different datasets with the trunk model VGG-16.}
	\label{fig:six}
\end{figure*}

\begin{figure}[t]
	\centering
	\includegraphics[width=0.35\textwidth]{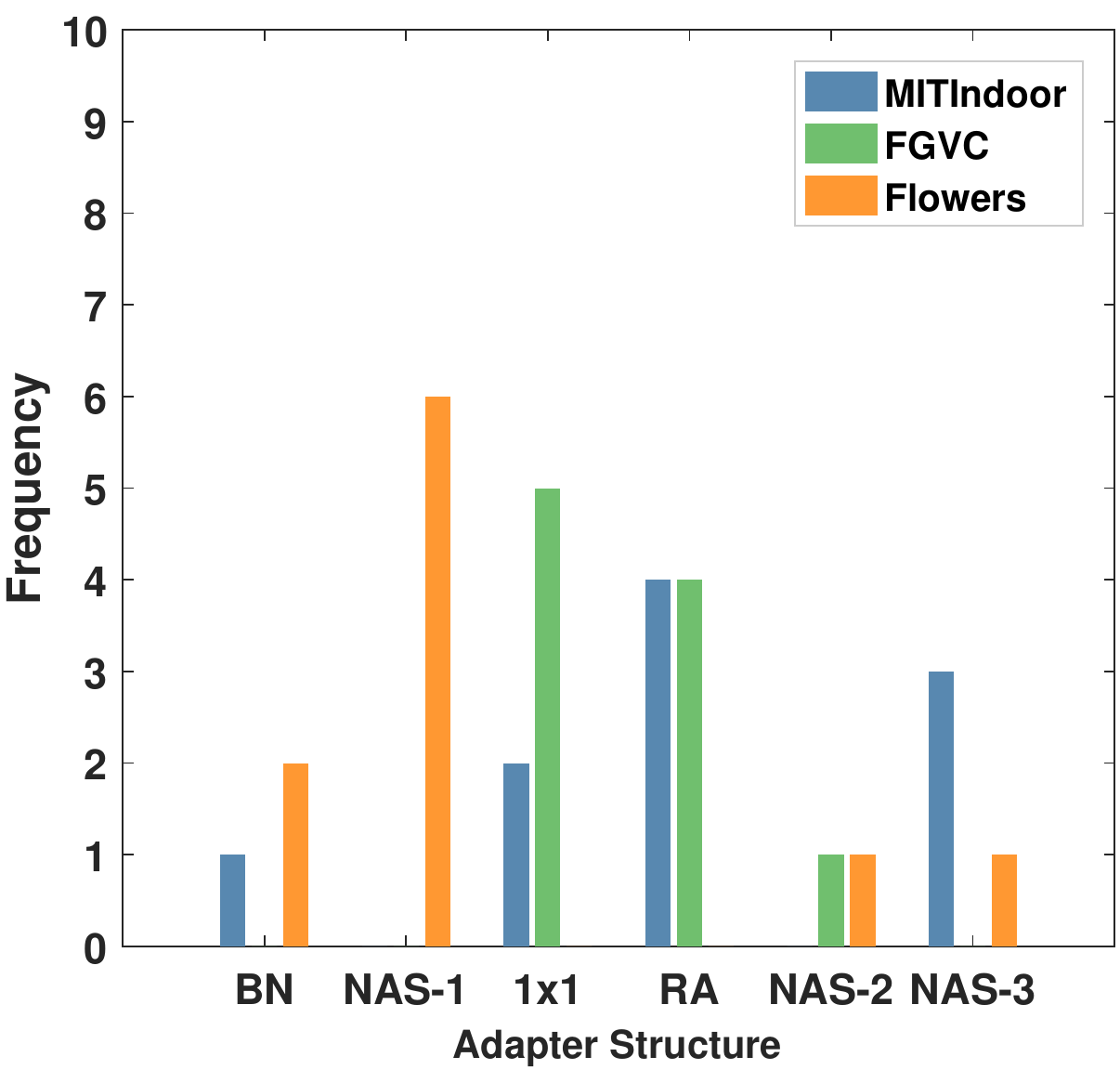}
	\caption{The distribution of learned adapter structure on different datasets with the trunk model VGG-16.}
	\label{rfig:one}
\end{figure}

\subsubsection{Effectiveness of our plugging strategy selection scheme} We evaluate our plugging strategy selection scheme with three kinds of baseline adapter structures. The plugging strategy controls whether or not to plug the adapter into a possible slot. For VGG-16, there are $15$ possible locations for plugging, while for ResNet-26 the number becomes $25$. 

Our plugging strategy selection scheme is capable of any hand-crafted adapter modules. $1 \times 1$ Adapt with the plugging strategy ``Ours" (Res Adapt with ``Ours" or BN Adapt-Part with ``Ours") denotes adding the $1 \times 1$ convolutional adapter modules (residual adapter modules or batch-normalize adapter modules) into the common trunk model at the locations decided by our selected plugging strategy. For the adapter parameter cost estimation, we use the plugging strategy ``All'' as a comparison point. We consider that ``All'' induce a $100\%$ parameter cost since they add adapters at all possible locations. It is then possible to calculate the relative cost for ``Ours".

The results of taking the VGG-16 as the common trunk model are presented in Table~\ref{tab:three}. It can be noticed that $1 \times 1$ Adapt with our selected plugging strategy (Res Adapt with ``Ours" or BN Adapt with ``Ours") has higher accuracy with $1 \times 1$ Adapt with the plugging strategy ``All" (Res Adapt with ``All" or BN Adapt-Part with ``All") but uses less additional parameters on three target datasets. This result demonstrates that some of the adapter modules perform a redundant role, which can be omitted without sacrificing accuracy. As for ResNet-26, reported in Table~\ref{tab:four}, with a different experiment setting (new trunk model and new datasets) we can observe that our optimized plugging strategy still consumes less extra parameters and obtains a competitive accuracy. This practice is especially important for VGG-16 since it contains a particularly large number of parameters, which has enough room for selecting an appropriate plugging strategy to reduce parameter cost of the adapters with accuracy improvement.

\subsubsection{Comparison with hand-crafted plugging strategies} In order to further demonstrate the effectiveness of our selected plugging strategy, we compare ``ours" with other three intuitively hand-crafted strategies. One of the intuitive plugging strategies is adding the adapter modules with a top-down order, i.e. select the first $n$ locations to plug the adapter modules. We denote this strategy as ``Top-Down''. Another intuitive plugging strategy is doing the opposite and adding the adapter modules with a bottom-up order, i.e. select the last $n$ locations to add the adapter, which is denoted as ``Bottom-Up''. A ``Random" strategy is also conducted by simply plugging the adapter module into $n$ random locations. For a fair comparison, we construct all the strategies by the same number of adapter modules. 

As shown in Figure~\ref{fig:four} and Figure~\ref{fig:five}, our strategy achieves the best performance on all datasets. For the Res Adapt module, the accuracies of the three hand-crafted strategies are nearly the same when using ResNet-26 as the trunk, while the Top-Down strategy fails with VGG-16. For the $1 \times 1$ Adapt, the three are also comparable with each other when using ResNet-26, but the Bottom-Up strategy occupies an inferior place with VGG-16. Also, with VGG-16 as the trunk, experiments on batch-normalize adapter module show that the accuracy of the Top-Down and Random plugging strategies are much higher than that of the Bottom-Up strategy. All these results demonstrate that the same plugging strategy performs differently depending on the adapter module structure and the trunk model, which fits our assumption.

We have also included a training time comparison for the entire pipeline in contrast to these hand-crafted plugging strategies (e.g. ``Random": randomly plugging adapter modules.
Compared to randomly plugging adapter modules, our method utilizes 40 hours for plugging strategy selection and improves the average accuracy by $2.27\%$ on the benchmark of seven domains.

\subsubsection{The importance of adapter structure search}
We evaluate our method with or without adapter structure search on three datasets. As shown in Table~\ref{tab:vgg:plugging}, the performance of our method with adapter structure search outperforms that without adapter structure search, which shows the importance of structure search to model performance.

\subsubsection{The distribution of learned adapter structure across domains}
We have conducted experiments to show the distribution of learned adapter structure across domains on three target datasets (i.e. MITIndoor, FGVC, Flowers) with the trunk model VGG-16 (For the domain complexity, MITIndoor $\textgreater$ FGVC $\textgreater$ Flowers) and the results are shown in Figure~\ref{rfig:one}. We search the adapter structure 10 times on each domain and calculate the frequency of each adapter structure. Six kinds of adapter structures are obtained in this experiment: BN Adapt, $1 \times 1$ Adapt, Res Adapt, NAS-1, NAS-2, and NAS-3. These adapter structures are listed in increasing order of complexity and shown in Figure~\ref{rfig:novel_module}. As shown in Figure~\ref{rfig:one}, the distribution of learned adapter structure for different domains varies. For the simple domain Flowers, NAS-1 and BN Adapt are more frequently selected. For the complex domain MITIndoor, our method tends to select Res Adapt and NAS-3. Simple adapter structures are usually selected on simple domains, and vice versa. All of these experiments demonstrate the diversity of our searching results on different domains, which shows that our method is effective.

\begin{figure*}[t]
	\centering
	\subfigure{}{
		\begin{minipage}[t]{1\textwidth}
			\includegraphics[width = 1\columnwidth]{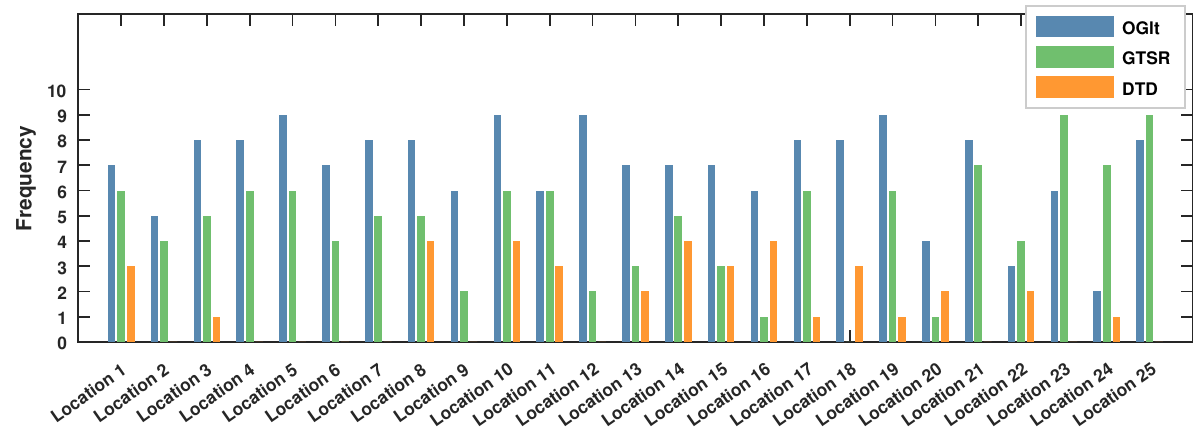}	
	\end{minipage}}
	\caption{Frequency of each plugging location to be selected on different datasets with the trunk model ResNet-26.}
	\label{fig:seven}
\end{figure*}

\subsubsection{Frequency of each plugging location to be selected} We show the frequency of each plugging location to be selected. For plugging strategy selection, we employ our method several times ($10$ in this experiment) and calculate the frequency of plugging an adapter at each plugging location. For VGG-16, there are $15$ possible plugging locations, while for ResNet-26 the number becomes $25$.
The results of taking the VGG-16 as the common trunk model are presented in Figure~\ref{fig:six}. On each dataset, the plugging locations with the highest frequency shows that these locations are often selected to plug an adapter, which implies they are important for learning this domain. For different datasets, the frequency of the same plugging location is different. All of these results demonstrate that a domain-specific plugging strategy is needed for each domain, which fits our motivation. Also, with ResNet-26 as the trunk in Figure~\ref{fig:seven}, we can obtain similar observations.

\begin{table*}[t]
	\centering
	\caption{Comparison of different paradigms on the benchmark of seven domains with the trunk model MobileNet. The best value is in \textbf{bold}.}
	\resizebox{1\textwidth}{!}{
		\begin{tabular}{lcccccccccc}
			\toprule
			Method  &FGVC &MITIndoor & Flowers & Caltech256 & SVHN & SUN397 &CIFAR100 & Ave. Acc.& Total Param.\\
			\midrule
			\midrule
			mobilenet-FNFT  &\textbf{79.63$\pm$0.14\%} &\textbf{68.94$\pm$0.08\%}& \textbf{95.69$\pm$0.11\%} & 82.71$\pm$0.17\% & \textbf{95.56$\pm$0.13\%} & 53.08$\pm$0.12\% &\textbf{78.90$\pm$0.08\%}& \textbf{79.22\%}&7\\
			\midrule
			mobilenet-Ours  & 79.55$\pm$0.05\%&68.43$\pm$0.11\%&94.51$\pm$0.05\%  &\textbf{84.09$\pm$0.03\%}  &95.42$\pm$0.02\%&\textbf{53.31$\pm$0.09\%}  & 78.84$\pm$0.11\%&79.16\% &\textbf{4.55}\\
			\bottomrule						
		\end{tabular}
	}
	\label{tab:mobilenet:sota:one}
\end{table*}	

\begin{table*}[t]
	\centering
	\caption{Accuracy, average accuracy, score and total parameter cost for seven popular vision datasets with the trunk model VGG-16. The best value is in \textbf{bold}.}
	\resizebox{1\textwidth}{!}{
		\begin{tabular}{lccccccccccc}
			\toprule
			Method  &FGVC &MITIndoor & Flowers & Caltech256 & SVHN & SUN397 &CIFAR100 & Ave. Acc. &S.& Total Param.&FLOP\\
			\midrule
			\midrule
			FNFT  &85.73\% &71.77\% & 95.67\% & 83.73\% & 96.41\% & 57.29\% &80.45\% & 81.58\%&1750&7& \textbf{1}\\
			\midrule
			\midrule
			BN~\cite{bilen2017universal}  &63.04\%& 57.60\% & 91.47\% & 73.66\% & 91.10\% & 47.04\% &64.80\% & 69.82\%&253& \textbf{$\approx$ 1}& $\approx$ 1 \\
			\midrule
			DAN~\cite{rosenfeld2018incremental}  &86.80\%& 63.02\% & 92.65\% & 68.63\% &  \textbf{96.55\%} & 45.98\% & 74.45\% & 75.44\%&957&2.84 &1.15\\	%
			\midrule										
			RA~\cite{rebuffi2017learning}  &88.92\% &72.40\%& 96.43\% & \textbf{84.17\%} & 96.13\% & \textbf{57.38\%}&79.55\%& 82.14\%&1935&2.85&1.15\\
			\midrule
			PA~\cite{rebuffi2018efficient}  &86.23\% &71.41\% & 95.20\% & 84.02\% & 96.05\% & 57.27\% &\textbf{79.85\%} & 81.43\%&1656&2.84&1.15\\
			\midrule
			BP-NAS~\cite{liu2020block}&89.01\% &72.53\% & 96.27\% & 83.64\% & 96.09\% & 57.14\% &79.36\% &82.01\%&1891&2.41&1.12\\
			\midrule
			PolSAR-DNAS~\cite{dong2020automatic}&86.59\% &70.13\% & 95.88\% & 83.48\% & 96.34\% & 57.26\% &78.59\% &81.18\%&1715&3.02&1.16\\
			\midrule
			Ours  & \textbf{89.34\%}&\textbf{73.51\%} &\textbf{96.96\%} &83.80\%  &96.47\%&57.28\%  & 79.48\%& \textbf{82.41\%}&\textbf{2082}&1.84&1.09\\
			\bottomrule						
		\end{tabular}
	}
	\label{tab:five}
\end{table*}

\subsubsection{The accuracy of our method with regard to domain diversity}
We give a detailed analysis about the accuracy of our method with regard to domain diversity. On the benchmark of seven domains, we construct two kinds of datasets: 1) the one contains the domains (i.e. FGVC+Flowers+SVHN covering particular fine-grained classes) which are more different from ImageNet (with general coarse-grained object classes); 2) the other one contains the domains (i.e. Caltech256+SUN197+CIFAR100) which are more similar to ImageNet. On these two datasets, we compare our method with the baseline RA~\cite{rebuffi2017learning}. As shown in Table~\ref{tab:five}, our method outperforms RA by a large margin on domains of the first dataset. The performance of the RA and our method are close to each other on the domains of the other dataset. From the results, we see that the performance gap increases on the dataset with more diverse domains.

\subsubsection{Comparison of different paradigms}
We compare two different paradigms on the benchmark of seven domains: one is training a smaller network from scratch for each domain (denoted by ``mobilenet-FNFT") and the other is employing our NAS-driven multi-domain learning method to plug a set of adapters to a trunk model MobileNet (denoted by ``mobilenet-Ours"). As shown in Table~\ref{tab:mobilenet:sota:one}, compared to mobilenet-FNFT, mobilenet-Ours achieves comparable performance and only utilizes about 65\% parameters.

\subsection{Comparison to Previous Methods}
In this section, we evaluate the MDL performance of our proposed scheme on two benchmarks, against other methods with different adapter structures and different architecture searching methods, including RA~\cite{rebuffi2017learning}, BN~\cite{bilen2017universal}, DAN~\cite{rosenfeld2018incremental}, PA~\cite{rebuffi2018efficient}, BP-NAS~\cite{liu2020block}, PolSAR-DNAS~\cite{dong2020automatic}.

\begin{table*}[t]
	\centering
	\caption{Accuracy, average accuracy, score and total parameter cost for the Visual Decathlon Challenge with the trunk model ResNet-26. The best value is in \textbf{bold}.}
	\resizebox{1\textwidth}{!}{
		\begin{tabular}{lccccccccccccc}
			\toprule
			Method  &ImNet &Airc. & C100 & DPed & DTD  &GTSR & Flwr &OGlt& SVHN&UCF& Ave. Acc. &S.& Total Param.\\
			\midrule
			\midrule
			FNFT  &59.87\% &60.34\% & 82.12\% & 92.82\% & 55.53\% & 97.53\% &81.41\% & 87.69\%&96.55\%&51.20\%& 76.51\% &2500& 10\\
			\midrule
			\midrule
			BN~\cite{bilen2017universal} & 59.87\% &43.05\% & 78.62\% & 92.07\% & 51.60\% & 95.82\% &74.14\% & 84.83\%&94.10\%&43.51\%& 71.76\% &1263& \textbf{$\approx$ 1}\\
			\midrule
			DAN~\cite{rosenfeld2018incremental}  &57.74\% &64.11\% & 80.07\% & 91.29\% & 56.54\% & 98.46\% &86.05\% & 89.67\%&\textbf{96.77\%}&49.38\%& 77.01\% &2851& 2.02\\
			\midrule									
			RA~\cite{rebuffi2017learning}&  59.23\% &63.73\% & 81.31\%& 93.30\% & 57.02\% & 97.47\% &83.43\% & 89.82\%&96.17\%&50.28\%& 77.17\% &2643& 2.03\\
			\midrule
			PA~\cite{rebuffi2018efficient}&  60.32\% &64.21\% & 81.91\% & \textbf{94.73\%} & 58.83\% & \textbf{99.38\%} &84.68\% & 89.21\%&96.54\%&\textbf{50.94\%}& 78.07\% &3412& 2.02\\
			\midrule
			BP-NAS~\cite{liu2020block}&  60.35\% &64.19\% & \textbf{81.92\%} & 94.67\%& 58.94\% & 98.77\% &84.64\% & 89.99\%&96.57\%&50.88\%& 78.09\% &3247& 1.86\\
			\midrule
			PolSAR-DNAS~\cite{dong2020automatic}&  59.97\% &64.14\% & 81.42\% & 93.54\% & 58.47\% & 98.34\% &83.96\% & 89.94\%&96.35\%&50.72\%& 77.69\% &2950& 2.31\\
			\midrule
			Ours & \textbf{60.43\%} &\textbf{64.32\% }& 81.70\% & 94.61\% & \textbf{59.47\%} &99.34\% & \textbf{84.77\%}& \textbf{90.02\%}&96.63\%&50.87\%& \textbf{78.22\%} &\textbf{3446}&1.54\\
			\bottomrule					
		\end{tabular}
	}
	\label{tab:six}
\end{table*}
\paragraph{Results on the benchmark of seven domains} 
We evaluate methods on the benchmark consisted of seven visual domains. The trunk structure we used is VGG-16. As shown in Table~\ref{tab:five}, FNFT, i.e. finetuning the full network for each domain, takes the most parameters since it uses a whole different version of the trunk for each domain. Our method yields $82.41\%$ average accuracy with $1.84$ times the number of parameters. Compared to RA, this accuracy rate is on par with their results, but the parameters cost is much lower (saving almost $55\%$ additional parameter cost). Although almost no additional parameters are added by BN, its average accuracy is $13\%$ lower than our method. As shown in Table~\ref{tab:five}, our method achieves better average accuracy and lower FLOPs than DAN. Compared to RA, PA, BP-NAS and PolSAR-NAS, our method also achieves lower FLOPs with comparable performance. The lower computational cost comes from the following aspects: 1) selecting an appropriate plugging strategy can reduce the computational cost because the number of adapters plugged to the trunk model is decreased; 2) searching a simple adapter structure on some domains can reduce the computational cost because the computational cost of an adapter is decreased. Our method only achieves comparable performance compared with the baseline because our method adaptively keeps a trade-off between the performance and memory cost for each domain. As shown in Table~\ref{tab:five}, the average accuracy of our method outperforms that of the baseline with a simple adapter structure BN by 12.59\%. Compared to the baseline with a complicated adapter structure RA, our method achieves comparable performance but only uses around 60\% memory cost. These results demonstrate that our method achieves a good balance between effectiveness and efficiency, which shows that our method is superior. We have also conducted experiments when enlarging the proposed model with similar total parameter cost to RA with the trunk model VGG-16. The average accuracy of our method with similar parameter cost to RA is increased by 0.64\% than before.

\paragraph{Results on the Visual Decathlon benchmark} 
We also analyze the performance on the Visual Decathlon benchmark, take the ResNet-26 as the trunk model. As shown in Table~\ref{tab:six}, BN utilizes the fewest parameters but has a poor performance across the tested domains. Both the average accuracy and score of our method is better than BN. Compared with other methods, our method achieves higher accuracy and score with less total parameter cost. On some specific domains such as DTD, our method owns the highest accuracy.

\section{Conclusion}\label{conclusion}
In this paper, we have proposed a novel NAS-driven multi-domain learning scheme, which aims to automatically set up the adapter plugging strategy and adaptively fulfill the adapter structure design. The proposed scheme is capable of utilizing NAS to learn where to plug as well as what adapter structure to plug. With the plugging strategy, our scheme is flexible in adapting to different domains. When compared to other methods, the MDL model obtained by our scheme is more compact and discriminative. Comprehensive experiments and analysis demonstrate the effectiveness of our scheme.

\section*{Acknowledgment}
This work is supported in part by National Key Research and Development Program of China under Grant 2020AAA0107400, National Natural Science Foundation of China under Grant U20A20222, Zhejiang Provincial Natural Science Foundation of China under Grant LR19F020004, and key scientific technological innovation research project by Ministry of Education

\bibliographystyle{IEEEtran}
\bibliography{ref}{}

\end{document}